\documentclass[letterpaper]{article} 
\usepackage{aaai2026}  
\usepackage{times}  
\usepackage{helvet}  
\usepackage{courier}  
\usepackage[hyphens]{url}  
\usepackage{graphicx} 
\usepackage{natbib}  
\usepackage{caption} 
\usepackage[ruled,vlined]{algorithm2e}

\usepackage{newfloat}
\usepackage{listings}
\usepackage{amsfonts}
\usepackage{amsmath, amssymb, amsthm}
\newtheorem{theorem}{Theorem}
\usepackage{xcolor}
\usepackage{makecell}
\usepackage{siunitx}
\usepackage{booktabs}
\usepackage{dsfont}
\newenvironment{proofsketch}{\begin{proof}[Proof sketch]}{\end{proof}}
\DeclareCaptionStyle{ruled}{labelfont=normalfont,labelsep=colon,strut=off} 
\title{Treatment Stitching with Schrödinger Bridge for Enhancing Offline Reinforcement Learning in Adaptive Treatment Strategies}
\author{
    Dong-Hee Shin, Deok-Joong Lee, Young-Han Son, Tae-Eui Kam
}
\affiliations{
    \textsuperscript{}Department of Artificial Intelligence, Korea University\\

    Seoul, Republic of Korea\\
    \{dongheeshin, deokjoong, yhson135, kamte\}@korea.ac.kr
}

\usepackage{bibentry}

\begin{document}

\maketitle

\begin{abstract}
Adaptive treatment strategies (ATS) are sequential decision-making processes that enable personalized care by dynamically adjusting treatment decisions in response to evolving patient symptoms. While reinforcement learning (RL) offers a promising approach for optimizing ATS, its conventional online trial-and-error learning mechanism is not permissible in clinical settings due to risks of harm to patients. Offline RL tackles this limitation by learning policies exclusively from historical treatment data, but its performance is often constrained by data scarcity—a pervasive challenge in clinical domains. To overcome this, we propose \textit{Treatment Stitching} (\textit{TreatStitch}), a novel data augmentation framework that generates clinically valid treatment trajectories by intelligently stitching segments from existing treatment data. Specifically, \textit{TreatStitch} identifies similar intermediate patient states across different trajectories and stitches their respective segments. Even when intermediate states are too dissimilar to stitch directly, \textit{TreatStitch} leverages the Schrödinger bridge method to generate smooth and energy-efficient bridging trajectories that connect dissimilar states. By augmenting these synthetic trajectories into the original dataset, offline RL can learn from a more diverse dataset, thereby improving its ability to optimize ATS. Extensive experiments across multiple treatment datasets demonstrate the effectiveness of \textit{TreatStitch} in enhancing offline RL performance. Furthermore, we provide a theoretical justification showing that \textit{TreatStitch} maintains clinical validity by avoiding out-of-distribution transitions.
\end{abstract}


\section{Introduction}
\begin{figure*}[t!]
    \centering
    \includegraphics[width=0.80\textwidth]{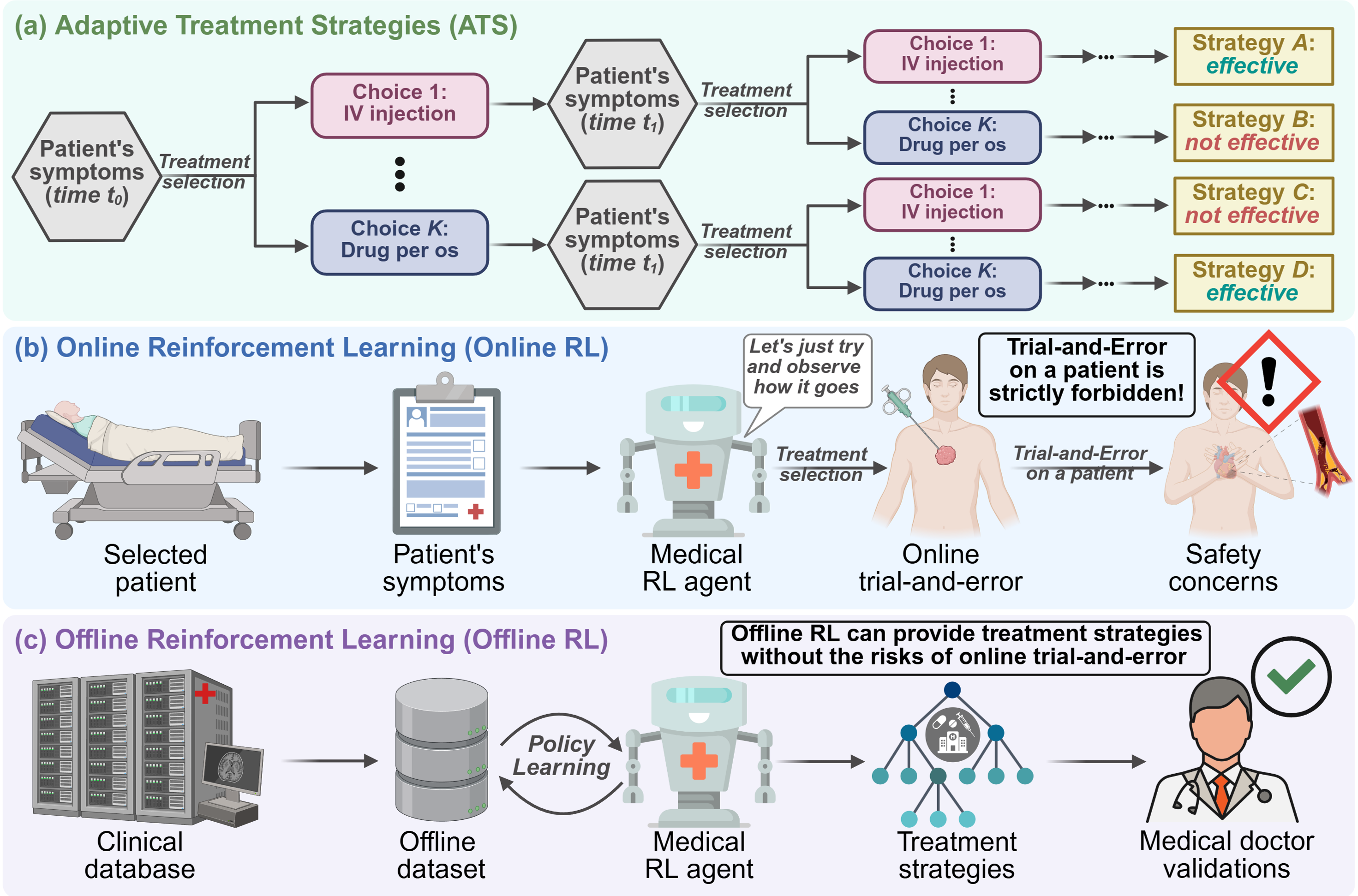}
    \caption{Illustrations of adaptive treatment strategies (ATS) and reinforcement learning (RL). (a) ATS aims to identify effective treatment strategies based on a patient’s evolving symptoms. (b) Online RL learns via trial-and-error, raising safety concerns in clinical settings. (c) Offline RL learns a policy from the offline dataset, removing the need for online trial-and-error on patients.}
    \label{fig1}
\end{figure*}

Imagine a scenario where a patient needs to visit the hospital regularly to manage chronic health diseases.  During each visit, clinicians assess the patient’s symptoms, review their medical history, and prescribe a treatment tailored to their current condition. After observing how the patient responds, clinicians utilize this new information to adapt treatment strategies to optimize patient outcomes. This dynamic process—where each decision is informed by a sequence of past observations, treatments, and responses—is referred to as an adaptive treatment strategy (ATS) \cite{ats}. As depicted in Figure \ref{fig1}(a), ATS play a crucial role in delivering personalized care in longitudinal clinical settings, where treatment decisions adapt to a patient's evolving symptoms.

In recent years, advancements in artificial intelligence (AI) have sparked a growing interest in developing AI-driven clinical decision support systems to enhance the implementation and optimization of ATS \cite{ai_dcss}. Early research in this domain has predominantly focused on behavior cloning (BC) \cite{bc_review}. To be specific, BC learns a policy through supervised learning, where the objective is to replicate clinicians' decisions at each time step by minimizing the discrepancy between AI-generated treatment recommendations and clinicians' actual decisions \cite{bc_treat}.

While BC is straightforward and intuitive, it prioritizes short-term alignment with clinicians' decisions rather than optimizing long-term clinical outcomes \cite{bc_limitations}. This limitation implies that BC can only reproduce past clinicians' decisions and may overlook potentially superior treatment strategies that could yield better long-term outcomes. To overcome this limitation, the reinforcement learning (RL) approach has been introduced as RL can optimize for long-term outcomes by maximizing cumulative rewards \cite{rl_ats,nips_dtr}. Moreover, RL enables an agent to actively explore its environment, allowing it to discover better solutions \cite{mars}.

However, a fundamental challenge in applying RL to clinical settings arises from its learning mechanism. As shown in Figure \ref{fig1}(b), conventional online RL learns a policy through an online trial-and-error process. In clinical settings, this would necessitate experimenting with different treatment strategies directly on patients to evaluate their efficacy, which raises significant ethical and safety concerns \cite{rl_healthcare}. To address this challenge, offline RL has emerged as a promising alternative \cite{offline_rl2,offline_rl1}. As shown in Figure \ref{fig1}(c), offline RL utilizes extensive clinical databases to construct a static offline dataset that contains historical treatment data collected from real-world clinical practices. This offline dataset is typically composed of sequences of states (symptoms), actions (prescribed treatments), and rewards (responses), providing a rich source of information for policy learning. Next, offline RL analyzes these historical data to identify treatment decisions that maximize cumulative rewards. As a result, offline RL can provide effective treatment strategies for clinicians without the risks of online trial-and-error experimentation on patients \cite{position_icml,aaai_ats}.

Despite the promise of offline RL, several challenges persist in its clinical applications. A well-known limitation of RL algorithms is their substantial data requirements for effective training \cite{rl_book}. This limitation implies a need for a large and comprehensive offline dataset to train robust offline RL policies. However, clinical settings are inherently data-hungry, often limited by the availability and diversity of historical treatment data \cite{data_hungry}. A common approach to address this data scarcity issue is data augmentation, particularly through the use of generative models, showing promising results in domains like medical imaging \cite{da_image1}. However, applying generative models to generate synthetic treatment data entirely from scratch presents unique challenges. Unlike medical images, treatment data often exhibit long-term causal dependencies across multiple treatment stages. Thus, generative models may struggle to capture these longitudinal dynamics, potentially leading to low-quality synthetic data. Moreover, generating synthetic data entirely from scratch can lead to error accumulation over time \cite{synthetic}.

In this paper, we introduce \textit{Treatment Stitching} (\textit{TreatStitch}), a novel data augmentation framework designed for offline RL in ATS applications. Unlike methods that rely on generative models to synthesize treatment data from scratch, \textit{TreatStitch} generates synthetic trajectories by intelligently stitching together segments from real patient trajectories in the offline dataset. Specifically, it identifies similar intermediate patient states (e.g., similar clinical conditions) across different trajectories. When such similar states are found, \textit{TreatStitch} `cuts' both trajectories at that point and `stitches' the first segment of one trajectory with the second segment of the other, creating a new, clinically valid stitched trajectory that preserves authentic state-action transitions.

However, when all trajectories do not share similar intermediate states, \textit{TreatStitch} leverages the Schrödinger bridge method \cite{schrodinger} to construct smooth and energy-efficient bridging trajectories between dissimilar states. This approach greatly expands data augmentation opportunities, particularly within sparse or heterogeneous offline datasets. Moreover, by restricting synthetic data generation to only these minimal bridging trajectories, we reduce the potential for error accumulation that can arise from generating extensive synthetic data from scratch. As a result, \textit{TreatStitch} significantly enhances the diversity and coverage of the offline dataset by augmenting these clinically valid stitched trajectories. The main contributions of this work are outlined as:

\begin{itemize}

\item To the best of our knowledge, \textit{TreatStitch} is the first data augmentation framework for ATS applications to utilize a trajectory stitching method that generates clinically valid synthetic trajectories from existing offline treatment data.

\item Even when offline data is sparse or heterogeneous, making direct stitching between states difficult, we introduce the Schrödinger bridge to construct smooth bridging trajectories that enable stitching between dissimilar states.

\item  We empirically demonstrate the effectiveness of \textit{TreatStitch} in enhancing offline RL performance and theoretically explain its capability to preserve clinical validity.

\end{itemize}

\clearpage
\section{Related Work and Preliminary}
\subsubsection{Adaptive Treatment Strategies (ATS).} ATS refer to clinical sequential decision-making systems that dynamically adapt treatment recommendations in response to evolving patient symptoms. Actions (i.e., treatment decisions) in ATS should take into account not only current symptoms, but also medical history and prior treatment responses, with the goal of optimizing long-term clinical outcomes rather than merely addressing immediate symptoms. In recent years, a diverse array of studies have been proposed to optimize ATS. Early methods utilized either online RL \cite{rl_ats} or imitation learning \cite{bc_treat} methods to tackle ATS optimization. However, these approaches have some limitations: online RL poses safety concerns due to online patient experimentation, while imitation learning merely replicates clinicians' past decisions without optimizing for long-term outcomes. To address these challenges, offline RL \cite{aaai_ats,jbhi_ats,aaai_ats1} has emerged as better and safer alternative method as it enables learning optimal policies from historical treatment data without requiring online experimentation on patients. However, existing offline RL methods primarily focus on algorithmic improvements while overlooking the critical data perspective—despite the fact that RL requires large amounts of diverse data for effective training \cite{offline_rl1}. Therefore, in this study, we introduce a data augmentation framework that generates clinically valid synthetic treatment data to enhance offline RL performance in ATS applications.
Extended related work section is provided in Appendix \textcolor{red}{\ref{extended_related_work}}.

\subsubsection{Markov Decision Process (MDP) Formulation in ATS.} RL problems are typically formulated as an MDP, which provides a mathematical framework for modeling the decision-making process. Formally, an MDP is defined by a tuple $\mathcal{M} = (\mathcal{S}, \mathcal{A}, \mathcal{F}, \mathcal{R}, \gamma)$, where $\mathcal{S}$ is the state space that represents the set of all possible patient symptoms; $\mathcal{A}$ is the action space, corresponding to all possible treatment decisions that clinicians can prescribe; $\mathcal{F}: \mathcal{S} \times \mathcal{A} \times \mathcal{S} \rightarrow [0,1]$ is the transition probability function that defines the probability of transitioning to next state; $\mathcal{R}: \mathcal{S} \times \mathcal{A} \to \mathbb{R}$ is the reward function that maps state-action pairs to scalar rewards, often representing patient responses or overall clinical outcomes; $\gamma \in [0,1]$ is the discount factor that determines the present value of future rewards. At each time step $t$, the RL agent receives the current state $s_t\in\mathcal{S}$ and executes an action $a_t\in\mathcal{A}$ according to a policy $\pi: \mathcal{S} \rightarrow \mathcal{A}$. This action leads to a next state $s_{t+1}$ and yields a reward $r_t = \mathcal{R}(s_t,a_t)$ that quantifies the clinical outcome. The main goal of the RL agent is to find an \textit{optimal policy} $\pi^*$ that maximizes the expected discounted cumulative reward such as follows:
\begin{equation}
\pi^* = \arg\max_{\pi} \mathbb{E}_{\tau \sim \pi} \left[\sum_{t=0}^{T} \gamma^t \mathcal{R}(s_t, a_t)\right],
\end{equation}
where $\tau = \{(s_0,a_0,r_0),\ldots,(s_T, a_T,r_T)\}$ denotes a treatment trajectory of length $T$. The typical online RL allows the agent to explore its environment to gather new trajectories and update the policy. However, in clinical settings, online exploration (i.e., trial-and-error on real patients) is strictly prohibited. This constraint motivates the use of offline RL.

\noindent
\subsubsection{Offline Reinforcement Learning (RL) in ATS.} In offline RL, the agent is prohibited from further interaction with the environment. Instead, the agent must learn from a static offline dataset $\mathcal{D}=\{\tau_1,\tau_2,\ldots,\tau_N\}$ of $N$ treatment trajectories collected from the historical clinical database. Thus, the main goal of offline RL is to effectively leverage the offline dataset to learn an optimal (or near-optimal) policy without further exploration. However, this offline setting introduces two key challenges: \textbf{(1)} \emph{distributional shift}, which occurs when the learned policy suffers from unreliable evaluation due to distribution mismatch; and \textbf{(2)} \emph{extrapolation error}, which results in overly optimistic predictions when dealing with out-of-distribution (OOD) state-action pairs \cite{offline_rl1}. To address these challenges, the Conservative Q-Learning (CQL) algorithm \cite{cql} has been proposed. Specifically, CQL introduces a conservative regularization that explicitly penalizes overly optimistic predictions by discouraging high Q-values on OOD (unseen) state-action pairs. The loss function for CQL can be expressed as:
\begin{equation}
\begin{aligned}
\mathcal{L}(\theta) \!=\!
&\underbrace{
    \mathbb{E}_{(s,a)\sim\mathcal{D}}
    \Big[\Big(Q_\theta(s,a)\!-\!\hat{\mathcal{B}}Q_\theta(s,a)\Big)^2\Big]
}_{\text{Bellman Error}} \\
&\quad + \beta\underbrace{
    \Big(\mathbb{E}_{a\sim\mu}\Big[Q_\theta(s,a)\Big]
    \! -\mathbb{E}_{a\sim\mathcal{D}}\Big[Q_\theta(s,a)\Big]\Big)
}_{\text{Conservative Regularization}},
\end{aligned}
\end{equation}
where $Q_\theta(s,a)$ represents the current Q-value estimate and $\hat{\mathcal{B}}$ is a Bellman backup operator \cite{rl_book} that refines this Q-value estimate by considering both the immediate reward and the discounted future reward. The first Bellman error term encourages $Q_\theta$ to accurately fit the observed trajectories in the offline dataset $\mathcal{D}$. The second regularization term penalizes $Q_\theta$ for OOD actions sampled from $\mu$, mitigating overly optimistic predictions. Note that $\mu$ typically represents the distribution for modeling OOD actions, such as a uniform distribution or the current policy $\pi$.

\begin{figure*}[t!]
    \centering
    \includegraphics[width=0.98\textwidth]{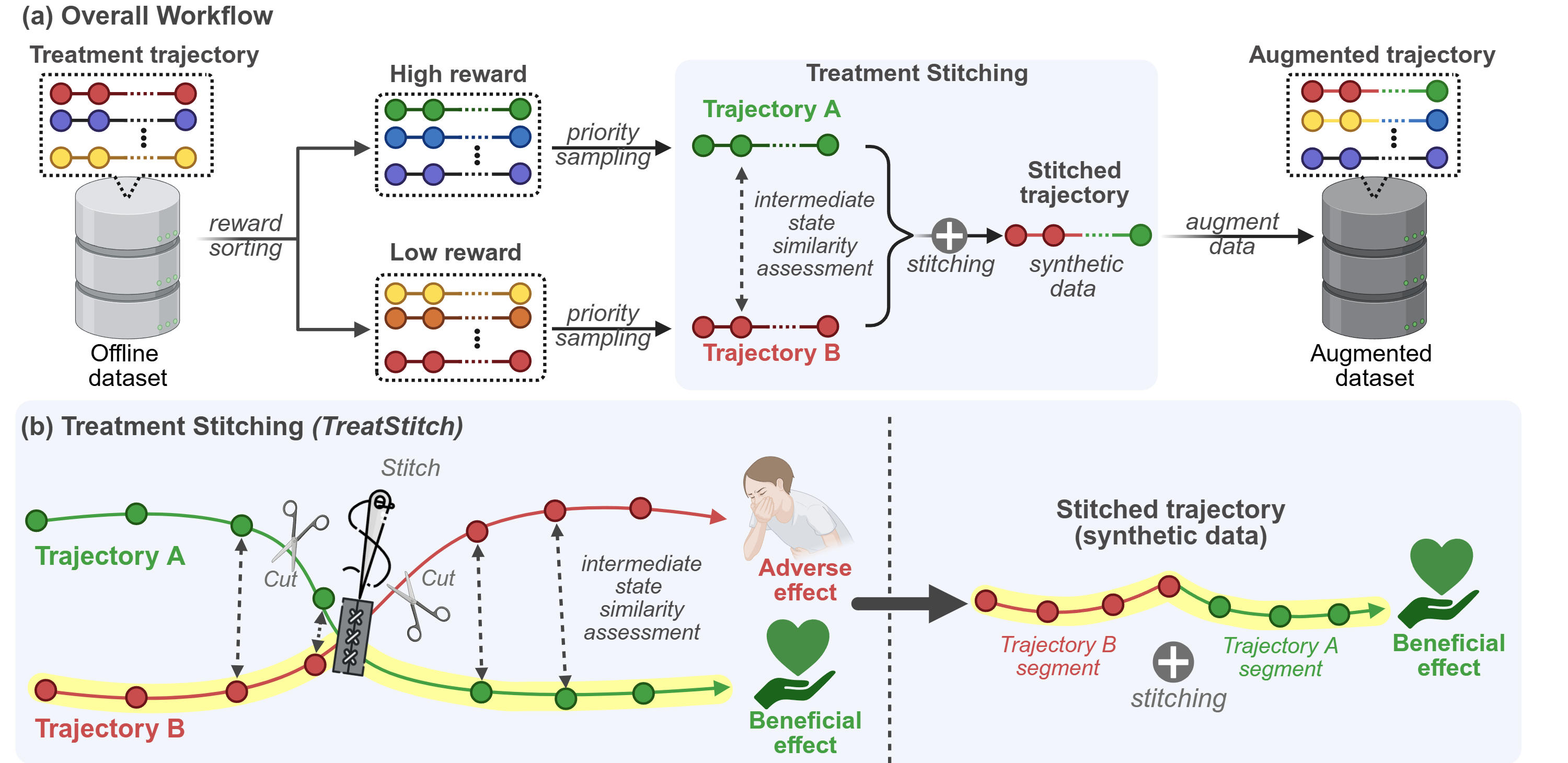}
    \caption{(a) The overall workflow of our treatment stitching framework that produces an augmented dataset for enhancing offline RL. (b) The detailed process of treatment stitching, which generates stitched trajectories from existing data.}
    \label{fig2}
\end{figure*}

\section{Method}
\subsection{Treatment Stitching Framework for Offline RL}
To address the data scarcity challenge in clinical settings, we propose the \textit{Treatment Stitching} (\textit{TreatStitch}) framework, which fully leverages existing treatment trajectories in offline datasets to generate new stitched treatment trajectories. As depicted in Figure \ref{fig2}(a), the overall process begins with the offline dataset $\mathcal{D} = \{\tau_1, \tau_2, \ldots, \tau_N\}$, where each trajectory $\tau_i = \{(s_t^i, a_t^i, r_t^i)\}_{t=0}^T $ consists of a sequence of states, actions, and rewards observed in clinical database. Then, the trajectories are categorized into two groups: the high reward group $\mathcal{D}_\text{high}$, consisting of trajectories that lead to beneficial clinical effects with high cumulative rewards, and the low reward group $\mathcal{D}_\text{low}$, which include trajectories with adverse effects and low cumulative rewards. Formally, it is given by:
\begin{equation*}
\mathcal{D}_{\text{high}} = \left\{ \tau_i \in \mathcal{D} \mid \sum_{t=0}^{T_i} \gamma^t r^i_t \geq \mathrm{\Phi}_q(\mathcal{D}) \right\}, 
\mathcal{D}_{\text{low}} = \mathcal{D} \setminus \mathcal{D}_{\text{high}},
\end{equation*}
where $\mathrm{\Phi}_q(\mathcal{D})$ denotes the reward value at the $q$-th percentile of cumulative rewards across all trajectories in $\mathcal{D}$. 

\subsubsection{Priority Sampling.}
Subsequently, we introduce priority sampling using the Boltzmann distribution to strategically select trajectories for the stitching process. The key insight is that combining trajectories with lower rewards (adverse effects) with those having higher rewards (beneficial effects) is more likely to produce informative stitched trajectories, where treatment trajectories beginning with adverse effects can still evolve into beneficial outcomes. To implement this intuition, our sampling strategy prioritizes trajectories with higher rewards in $\mathcal{D}_{\text{high}}$ and lower rewards in $\mathcal{D}_{\text{low}}$. Formally, the sampling probability of trajectory $\tau_i$ in $\mathcal{D}_{\text{high}}$ is given by:
\begin{equation}
p(\tau_i \mid \tau_i \in \mathcal{D}_{\text{high}}) 
= \frac{\exp(R(\tau_i)/\alpha)}{\sum_{\tau_j \in \mathcal{D}_{\text{high}}} \exp(R(\tau_j)/\alpha)},
\end{equation}
where $R(\tau_i) = \sum_{t=0}^{T} \gamma^t r^i_t$ is the cumulative reward of $\tau_i$, and $\alpha$ is the temperature parameter. For trajectories in $\mathcal{D}_{\text{low}}$, we use $-R(\tau_i)$ to prioritize trajectories with lower rewards. As training progresses, we gradually increase $\alpha$ to shift from highly prioritized to uniform sampling for broader coverage. Further analysis of priority sampling is in Appendix \textcolor{red}{\ref{appendix_priority_sampling}}.

\subsubsection{Treatment Stitching.} Trajectory A ($\tau_A$) and Trajectory B ($\tau_B$) are sampled from $\mathcal{D}_{\text{high}}$ and $\mathcal{D}_{\text{low}}$, respectively, using priority sampling. To identify a potential stitching point, we perform an \textit{intermediate state similarity assessment} by computing the cosine similarity between all pairs of intermediate states—comparing each state $\{s^A_t\}$ in $\tau_A$ with $\{s^B_{t'}\}$ in $\tau_B$:
\begin{equation}
\label{eq:cosine_similarity_t}
\text{Sim}\left(s_t^A, s_{t'}^B\right) = \frac{\langle s_t^A, s_{t'}^B \rangle}{\|s_t^A\| \|s_{t'}^B\|}.
\end{equation}
If $\text{Sim}\left(s_t^A, s_{t'}^B\right) \geq \delta$, where $\delta$ is a predefined threshold, this pair of states is selected as a stitching point. Otherwise, new trajectories $\tau_A$ and $\tau_B$ are sampled, and the process repeats.

Once a valid stitching point is identified, two trajectories $\tau_A$ and $\tau_B$ are combined to construct a stitched trajectory $\tau_{\text{stit}}$. We concatenate the segments of $\tau_B$ from time 0 through $t'$ and the segments of $\tau_A$ from $t+1$ to its terminal point $T$:
\begin{equation}
\begin{aligned}
\tau_{\text{stit}} = 
&\left\langle
\left(s_0^B, a_0^B, r_0^B\right), \dots, \left(s_{t'}^B, a_{t'}^B, r_{t'}^B\right)
\right\rangle \\
&\oplus\,
\left\langle
\left(s_{t+1}^A, a_{t+1}^A, r_{t+1}^A\right), \dots, \left(s_T^A, a_T^A, r_T^A\right)
\right\rangle.
\end{aligned}
\label{stitching_equation}
\end{equation}
The newly created stitched trajectories $\tau_{\text{stit}}$ are incorporated into the original offline dataset $\mathcal{D}$ to create an augmented dataset $\mathcal{D}_{\text{aug}}$. Finally, we then use $\mathcal{D}_{\text{aug}}$ to train the offline RL algorithms, as its increased diversity and broader coverage of treatment trajectories lead to improved performance.

As depicted in Figure \ref{fig2}(b), our \textit{treatment stitching} process effectively combines beneficial segments from different trajectories. For instance, even though Trajectory B culminates in an adverse effect, its initial to intermediate transitions can still reflect meaningful clinical behaviors. By identifying a high-similarity state pair and merging the early segment of Trajectory B with the later segment of Trajectory A, we generate a new `stitched trajectory' that inherits favorable characteristics from both. This enables the creation of clinically valid treatment trajectories that are not directly present in $\mathcal{D}$.

Despite its effectiveness, this direct stitching relies on the assumption that the offline dataset $\mathcal{D}$ is sufficiently large or homogeneous to contain enough similar intermediate states across trajectories for identifying valid stitching points. Unfortunately, in clinical practice, this assumption may not always hold. Clinical offline datasets can be extremely sparse or heterogeneous, making it challenging—or even impossible—to find valid stitching points due to the lack of similar intermediate states between trajectories. As a result, direct stitching alone may be inadequate in certain clinical settings.

\clearpage

\begin{figure}[t]
\includegraphics[width=0.9\columnwidth]{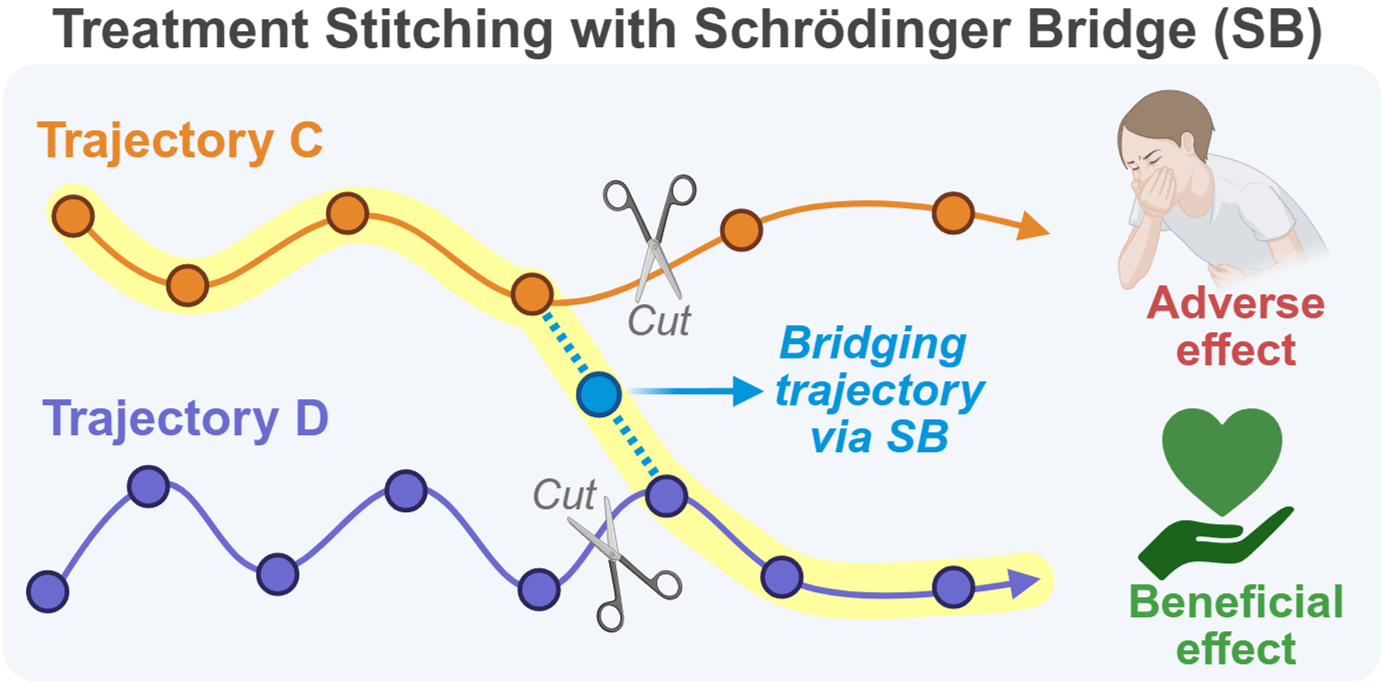}
\centering
\caption{Overview of Schrödinger bridge for \textit{TreatStitch}.}  
\label{sb_figure}
\end{figure} 
\subsection{Schrödinger Bridge for Treatment Stitching}
To tackle this challenge, we further introduce a novel mechanism that leverages the Schrödinger bridge method \cite{schrodinger} to construct smooth and energy-efficient bridging trajectories that enable stitching even between dissimilar states. This enhancement broadens the applicability of our \textit{TreatStitch} to sparse and heterogeneous clinical datasets.

\subsubsection{Schrödinger Bridge.} 
Originally, Erwin Schrödinger studied the problem of identifying the most probable and smooth stochastic trajectory that connects two given probability distributions \cite{schrodinger_original}. Nowadays, this is commonly referred to as the \textit{Schrödinger bridge (SB) problem}, which is rooted in optimal transport (OT) theory \cite{ot}.

Formally, given a starting distribution $p_{\text{start}}$ and a target distribution $p_{\text{target}}$, and a reference stochastic process (e.g., Brownian motion), the goal is to find an optimal stochastic process $\mathbb{P}^*$ that transports samples from $p_{\text{start}}$ to $p_{\text{target}}$ over a finite time interval. This optimality is defined by minimizing the Kullback–Leibler (KL) divergence between the path measure of the target process $\mathbb{P}$ and the reference process $\mathbb{Q}$:
\begin{equation}
\mathbb{P}^* = \underset{\mathbb{P}}{\arg\min} \, \mathrm{KL}(\mathbb{P} \| \mathbb{Q}) \; \text{s.t.} \; \mathbb{P}_0 = p_{\text{start}}, \mathbb{P}_T = p_{\text{target}}.
\label{eq:sbp}
\end{equation}
A key result is that the solution to this problem is characterized by the Schrödinger system \cite{caluya}, which consists of a pair of partial differential equations:
\begin{equation}
\begin{cases}
\frac{\partial \Psi}{\partial t} = -\langle \nabla \Psi, f \rangle - \frac{1}{2} g^2 \Delta \Psi \\
\frac{\partial \hat{\Psi}}{\partial t} = -\nabla \cdot [\hat{\Psi} f] + \frac{1}{2} g^2 \Delta \hat{\Psi}
\end{cases}
\hspace{-3mm} \text{s.t.} \hspace{-0.5mm}
\begin{cases}
\Psi(\cdot, 0)\hat{\Psi}(\cdot, 0)\!=\!p_{\text{start}}, \\
\Psi(\cdot, T)\hat{\Psi}(\cdot, T)\!=\!p_{\text{target}},
\end{cases}
\label{schrodinger_system}
\end{equation}
where $\Psi$ and $\hat{\Psi}$ are the Schrödinger potentials, $f$ denotes the drift vector field, and $g$ is the scalar diffusion coefficient. A more detailed analysis of the SB problem is in Appendix \textcolor{red}{\ref{extended_SB}}.

\subsubsection{Problem Formulation.} In terms of \textit{TreatStitch}, we formulate the construction of bridging trajectories between dissimilar intermediate states as the SB problem. As shown in Figure \ref{sb_figure}, suppose we have two trajectories, Trajectory C ($\tau_C$) and Trajectory D ($\tau_D$), where no pair of intermediate states satisfies the similarity threshold $\delta$. Our goal is to construct a bridging trajectory $\tau_{\text{bridge}}$ that connects a selected pair of intermediate states—$s_{t}^C \in \tau_C$ and $s_{t'}^D \in \tau_D$—identified as the most similar among all candidates despite still falling below the similarity threshold $\delta$. This bridging trajectory $\tau_{\text{bridge}}$ serves to facilitate the stitching process between $\tau_C$ and $\tau_D$.

\subsubsection{Bridging State Generation.}
We first generate bridging states using the SB method. Specifically, we construct a neural network $G_\phi$ that learns the OT-based stochastic trajectory connecting two given probability distributions by solving the Schrödinger system as stated in Equation \ref{schrodinger_system}. Given the start state $s_{t}^C$ and the target state $s_{t'}^D$, the goal of $G_\phi$ is to generate a sequence $s_{\text{bridge}} = \{ \tilde{s}_1, \tilde{s}_2, \dots, \tilde{s}_K \}$ that ensures a smooth transition from $s_{t}^C$ to $s_{t'}^D$. Moreover, the number of bridging states $K$ is minimized to mitigate the risk of error accumulation resulting from extensive synthetic data generation. Due to space constraints, the detailed training objective and the step-by-step procedure for $G_\phi$ are provided in Appendix~\textcolor{red}{\ref{appendix_obj}}.

\subsubsection{Bridging Trajectory Completion.}
Once the sequence of $s_{\text{bridge}}$ is obtained, we complete the bridging trajectory $\tau_{\text{bridge}}$ by inferring the corresponding actions and rewards for each transition. To this end, we construct two neural networks: an inverse dynamics model $I_\psi(a_t|s_t,s_{t+1})$ that predicts the action given a pair of successive states, and a reward prediction model $R_\rho(r_t|s_t, a_t)$ that estimates the reward based on a given state-action pair \cite{bitrajdiff}. These models are trained jointly using a supervised learning objective over trajectories sampled from $\mathcal{D}$. The combined loss function is:
\begin{equation}
\mathcal{L}(\psi, \rho) = \mathbb{E}_{{\scriptscriptstyle \tau \sim \mathcal{D}}} \hspace{-1mm} \left[ 
\left\| I_\psi(s_t, s_{t+1})\hspace{-0.5mm} - \hspace{-0.5mm}a_t \right\|^2 \hspace{-0.5mm} + \hspace{-0.1mm}
\left\| R_\rho(s_t, a_t) \hspace{-0.5mm}-\hspace{-0.5mm} r_t \right\|^2 
\right].
\end{equation}
After training, we apply $I_\psi$ and $R_\rho$ to each adjacent pair of bridge states $(\tilde{s}_i, \tilde{s}_{i+1})$ to generate the corresponding action $\tilde{a}_i$ and reward $\tilde{r}_i$, thereby forming the bridging trajectory:
\begin{equation}
\tau_{\text{bridge}} = \left\langle(\tilde{s}_1,\! \tilde{a}_1,\! \tilde{r}_1),\! (\tilde{s}_2,\! \tilde{a}_2,\! \tilde{r}_2),\! \dots,\! (\tilde{s}_K,\! \tilde{a}_K,\! \tilde{r}_K)\right\rangle.
\end{equation}
\subsubsection{Stitched Trajectory via Schrödinger Bridge}
Finally, we can construct an additional stitched trajectory via SB method $\tau_{\text{stit}}^{\text{SB}}$ by concatenating the initial segment of $\tau_C$ up to state $s_t^C$, the bridging trajectory $\tau_{\text{bridge}}$, and the later segment of $\tau_D$:
\begin{equation}
\begin{aligned}
\tau_{\text{stit}}^{\text{SB}} =
&\left\langle (s_0^C, a_0^C, r_0^C), \dots, (s_t^C, a_t^C, r_t^C) \right\rangle 
\oplus\, \tau_{\text{bridge}} \\
&\oplus\, \left\langle (s_{t'}^D, a_{t'}^D, r_{t'}^D), \dots, (s_T^D, a_T^D, r_T^D) \right\rangle
\end{aligned}
\label{sb_stitching_eq}
\end{equation}

\subsection{Theoretical Justification for Treatment Stitching}
We present the theoretical justification for the advantages of \textit{TreatStitch}, showing that it preserves clinical validity by mitigating distribution shifts and avoiding OOD transitions.
\begin{theorem}
Let $\mathcal{F}: \mathcal{S} \times \mathcal{A} \to \mathcal{S}$ be the transition function defining the environment dynamics, and fix a norm $\|\cdot\|$ on $\mathcal{S}$. Suppose $\mathcal{F}$ is $L$-Lipschitz continuous in the state coordinate:
\begin{equation}
\|\mathcal{F}(s, a) - \mathcal{F}(s', a)\| \leq L \|s - s'\| \hspace{2mm}
  \forall\,s,s'\in \mathcal{S}, a\in \mathcal{A}.
\end{equation}
Given the offline dataset $\mathcal{D} = \{\tau_i\}_{i=1}^N$, we construct the stitched trajectory $\tau_{\text{stit}}= \tau_B[0:t'] \cup \tau_A[t+1:T]$, where the stitching point is chosen to satisfy a similarity condition, allowing a maximum difference of $1 - \delta$. Then, for every transition \((\bar{s}, a, \bar{s}')\) in $\tau_{\text{stit}}$, there exists at least one transition \((s, a, s')\) in $\mathcal{D}$ such that:
\begin{equation}
    \|\bar{s} - s\| \leq \sqrt{2(1 - \delta)} \; \land \; \|\bar{s}' - s'\| \leq L \sqrt{2(1 - \delta)}.
\end{equation}
Consequently, $\tau_{\text{stit}}$ remains within an $\mathcal{O}\big(L \sqrt{2(1 - \delta)}\big)$-neighborhood of transitions already in $\mathcal{D}$. In other words, our treatment stitching process 
stays within a small neighborhood of the original data support, reducing OOD shifts.
\end{theorem}

\newpage
\begin{table}[t]
\scriptsize
\centering
\renewcommand{\arraystretch}{1.2} 
\setlength{\tabcolsep}{1.9pt} 
\begin{tabular}{c|cccccccc|c}
\noalign{\smallskip}\noalign{\smallskip} \hline\hline
\textbf{Method}   & \textbf{Env1}   & \textbf{Env2}   & \textbf{Env3}   & \textbf{Env4}   & \textbf{Env5}   & \textbf{Env6}   & \textbf{Env7}   & \textbf{Env8}    & \textbf{Mean}    \\
\cline{1-10}
\multicolumn{1}{l|}{CQL (Backbone)} & 59.65 & 57.03   & 47.25  & 53.82  & 54.35   & 54.68  & 55.59  & 52.55  & 54.37  \\
\multicolumn{1}{l|}{+ GAN} & 32.99  & 50.20   & 38.72  & 29.20  & 16.82   & 2.99  & 29.48  & 20.09  & 27.56 \\
\multicolumn{1}{l|}{+ DDPM} & 33.68  & 26.74   & 45.40  & 49.19  & 38.05   & 42.44  & 25.87  & 46.88  & 38.53 \\
\multicolumn{1}{l|}{+ SynthER} & 33.09  & 25.90   & 40.74  & 31.29  & 44.62   & 53.25  & 28.84  & 49.83  & 38.45 \\
\multicolumn{1}{l|}{+ GTA} & 61.75  & 56.55   & 43.26  & 53.17  & 47.60   & 56.96  & 57.55  & 47.82  & 53.08 \\
\multicolumn{1}{l|}{+ ATraDiff} & 61.66  & 54.38   & 49.31  & 52.63  & 50.47   & 58.10  & 52.76  & 48.34  & 53.46 \\
\multicolumn{1}{l|}{+ RTDiff} & 63.94  & 55.43   & 47.81  & 56.89  & 51.86   & 56.78  & 58.03  & 49.52  & 55.03 \\
\cline{1-10}
\multicolumn{1}{l|}{\textbf{+ TreatStitch}} & 65.15 & \textbf{62.05}  & 50.57  & 58.22  & \textbf{57.88}  & 56.98  & 59.09  & \textbf{53.51}  & 57.93  \\
\multicolumn{1}{l|}{\textbf{+ TreatStitch w/ SB}} & \textbf{66.76} & 60.32  & \textbf{51.76}  & \textbf{58.88}  & 55.83  & \textbf{58.33}  & \textbf{60.54}  & 52.03  & \textbf{58.06}  \\

\hline \hline
\end{tabular}
\caption{Performance comparison of each method using the CQL backbone on EpiCare under the \textbf{full data} setting.}
\label{main_full}
\end{table}

\section{Experimental Results and Discussion}
\subsubsection{Datasets.} To evaluate the performance of our \textit{TreatStitch}, we utilized the EpiCare benchmark \cite{epicare}, which is one of the most recent and comprehensive benchmarks for medical treatment evaluation. In particular, we selected this benchmark for our main experiments due to its appeal as a generalized benchmark, which makes it well-suited for longitudinal clinical settings without being limited to a specific disease. The EpiCare benchmark consists of datasets containing $2^{17}=131,072$ episodes for each of 8 distinct environment settings. As described in the benchmark, these datasets can be interpreted as being drawn from sequential treatment trials for 8 distinct diseases. In addition, we also evaluated our framework on the MIMIC-III database \cite{mimic}, which contains real-world electronic health record datasets. Specifically, we followed the experimental setup of prior work \cite{position_icml} that focuses on sepsis treatment in the intensive care unit (ICU).

\subsubsection{Setup.} We closely followed the experimental protocol outlined in the EpiCare benchmark for our main experiments. Specifically, we used the average cumulative reward as the evaluation metric. For reward design, we assigned $+64$ for remission, $-64$ for adverse events, and a penalty in the range of $[-1, -4]$ to reflect treatment costs. We set the similarity threshold $\delta = 0.95$  to ensure clinically valid stitching while maintaining sufficient synthetic samples. For the MIMIC-III dataset, we also followed the experimental setup established by prior work \cite{position_icml}. We evaluated performance using root mean squared error (MSE) for both IV fluid treatment (MSE\textsubscript{iv}) and vasopressor treatment (MSE\textsubscript{va}), along with weighted importance sampling (WIS) and doubly robust (DR). In terms of reward design, the sequential organ failure assessment (SOFA) score \cite{sofa} was included to measure the severity of organ dysfunction, and the national early warning score 2 (NEWS2) \cite{news} was used to estimate the risk of mortality. Further details are provided in Appendix~\textcolor{red}{\ref{appendix_setting}}.

\subsubsection{Competing methods.} In this study, we used CQL \cite{cql} as the main backbone offline RL algorithm due to its superior performance compared to other methods (see Appendix \textcolor{red}{\ref{appendix_backbone}}) and its widespread adoption as a backbone in various prior works \cite{aaai_ats,jbhi_ats}. Since our \textit{TreatStitch} is a data augmentation framework, we compared it against other data augmentation methods in RL domain. First, we considered established generative models, including generative adversarial network (GAN) \cite{gan} and diffusion models such as DDPM \cite{ddpm}. Specifically, we trained GAN and DDPM to generate synthetic treatment trajectories from scratch, which were then employed to augment the original dataset. Next, we incorporated SynthER \cite{synther}, which leverages diffusion models to augment synthetic data and enhances training through synthetic experience replay. We also included GTA \cite{gta}, which extends SynthER by augmenting synthetic trajectories to be both high-reward and dynamically consistent. Additionally, we evaluated ATraDiff \cite{atradiff}, which uses a coarse-to-fine strategy to efficiently generate synthetic trajectories. Finally, we included RTDiff \cite{rtdiff}, which is a state-of-the-art method that synthesizes trajectories in the reverse direction for more effective data augmentation.

\begin{table}[t]
\scriptsize
\centering
\renewcommand{\arraystretch}{1.2} 
\setlength{\tabcolsep}{1.9pt} 
\begin{tabular}{c|cccccccc|c}
\noalign{\smallskip}\noalign{\smallskip} \hline\hline
\textbf{Method}   & \textbf{Env1}   & \textbf{Env2}   & \textbf{Env3}   & \textbf{Env4}   & \textbf{Env5}   & \textbf{Env6}   & \textbf{Env7}   & \textbf{Env8}    & \textbf{Mean}    \\
\cline{1-10}
\multicolumn{1}{l|}{CQL (Backbone)} & 20.14 & 18.48 & 12.99  & 16.18 & 9.03 & 20.26 & 10.63 & 18.60 & 15.79   \\
\multicolumn{1}{l|}{+ GAN}  & 21.86 & 14.72  & 1.14 & 10.84 & 12.19 & 6.43 & 2.57 & 6.72 & 9.56 \\
\multicolumn{1}{l|}{+ DDPM}  & 13.53 & 15.48  & 17.18 & 9.99 & 16.94 & 12.95 & 5.60 & 5.17 & 12.11 \\
\multicolumn{1}{l|}{+ SynthER}  & 11.13 & 23.43  & 13.16 & 6.14 & 12.67 & 16.53 & 4.36 & 14.19 & 12.70 \\
\multicolumn{1}{l|}{+ GTA} & 19.93 & 20.83 & 18.27  & 24.21& 12.23 & 20.72 & 8.52 & 18.57 & 17.91   \\
\multicolumn{1}{l|}{+ ATraDiff}  & 13.29 & 21.53  & 14.24 & 24.49 & 15.53 & 25.10 & 13.65 & 13.43 & 17.66 \\
\multicolumn{1}{l|}{+ RTDiff} & 23.27 & 20.91  & 17.35 & 25.38 & 14.80 & 21.47 & 12.93 & 24.57 & 20.09 \\
\cline{1-10}
\multicolumn{1}{l|}{\textbf{+ TreatStitch}} & 42.22 & 43.21  & 30.29  & 33.63  & 34.78  & 34.68  & 36.03  & 30.19 & 35.63  \\
\multicolumn{1}{l|}{\textbf{+ TreatStitch w/ SB}} & \textbf{48.47} & \textbf{47.36}  & \textbf{34.00} & \textbf{36.94}  & \textbf{37.10}  & \textbf{36.65}  & \textbf{41.43}  & \textbf{38.64}  & \textbf{40.07}  \\

\hline \hline
\end{tabular}
\caption{Performance comparison of each method using the CQL backbone on EpiCare under the \textbf{restricted data} setting}
\label{main_restricted}
\end{table}

\subsubsection{Full Data Results.} 
As shown in Table \ref{main_full}, we evaluated the performance of each method under the full data setting using 131,072 episodes. Both GAN and DDPM significantly underperformed compared to the backbone, highlighting the risks of generating synthetic trajectories from scratch. This drop in performance is consistent with findings from a prior study \cite{nature} that describe model collapse, where generative models produce erroneous and low-quality synthetic data that leads to performance decline. Among the competing methods, RTDiff was the only method to show a performance improvement. However, this gain was limited, because RTDiff was designed for general RL tasks without considering clinical validity. In contrast, our \textit{TreatStitch} demonstrated superior performance by leveraging the treatment stitching that preserves clinical validity. Rather than generating synthetic data from scratch, \textit{TreatStitch} creates clinically valid new trajectories by intelligently combining segments from existing trajectories. The extended version, \textit{TreatStitch w/ SB}, achieved slightly better performance than \textit{TreatStitch}. However, this improvement was marginal, likely because the full data setting already provided abundant valid stitching points, thereby limiting additional gains from SB.

\begin{table}[t]
\scriptsize
\centering
\renewcommand{\arraystretch}{1.2}
\setlength{\tabcolsep}{1.9pt}
\begin{tabular}{c|cccc|cccc}
\hline\hline
\textbf{Method} & \multicolumn{4}{c|}{\textbf{SOFA ($\downarrow$)}} & \multicolumn{4}{c}{\textbf{NEWS2 ($\downarrow$)}} \\
\cline{2-9}
& {\textbf{MSE\textsubscript{iv}}} & {\textbf{MSE\textsubscript{va}}} & {\textbf{WIS}} & {\textbf{DR}} 
& {\textbf{MSE\textsubscript{iv}}} & {\textbf{MSE\textsubscript{va}}} & {\textbf{WIS}} & {\textbf{DR}} \\
\hline
\multicolumn{1}{l|}{CQL {(Backbone)}} & 611.30 & 0.39 & 13.27 & -0.37 & 566.64 & 0.33 & -4.70 & -0.69 \\
\multicolumn{1}{l|}{+ GTA} & 634.81 & 0.42 & 14.45 & -0.39  &  574.20 & 0.37 & -5.21 & -0.73  \\
\multicolumn{1}{l|}{+ ATraDiff} & 607.75 & 0.39 & 13.53  & -0.37 & 552.32  & 0.32 & -4.54 & -0.66 \\
\multicolumn{1}{l|}{+ RTDiff}  & 601.77  & 0.32 & 12.92 & -0.34 & 550.06 & 0.28 &  -4.13 & -0.62 \\
\cline{1-9}
\multicolumn{1}{l|}{\textbf{+ TreatStitch}} & 589.59 & \textbf{0.28} & 11.10  & -0.31 & 529.48 & 0.25 & -3.60 & \textbf{-0.56} \\
\multicolumn{1}{l|}{\textbf{+ TreatStitch w/ SB}} & \textbf{578.90} & \textbf{0.28} & \textbf{10.86}  & \textbf{-0.30} & \textbf{526.71} & \textbf{0.24} & \textbf{-3.33} & \textbf{-0.56} \\
\hline\hline
\end{tabular}
\caption{Performance comparison of each method using the CQL backbone on the real-world \textbf{MIMIC-III} dataset.}
\label{mimic_table}
\end{table}

\subsubsection{Restricted Data Results.} 
Since certain clinical settings often suffer from limited data availability, we conducted additional experiments under the restricted data setting, using only $2^{10}=1024$ episodes—128 times fewer than in the full data setting. This setting reflects scenarios where only limited patient treatment data is available, resulting in a sparse offline dataset. As shown in Table \ref{main_restricted}, both GAN and DDPM again demonstrated poor performance, confirming their fundamental limitations. Interestingly, other data augmentation methods—including GTA, ATraDiff, and RTDiff—showed modest performance improvements compared to the backbone. This contrasts with their performance in the full data setting and suggests that when data is extremely sparse, RL-specific augmentations methods can still offer some benefit by diversifying the training distribution. However, these improvements still remain limited, as these methods do not account for clinical validity in their design. In contrast, our \textit{TreatStitch} achieved significantly better performance by preserving clinical validity. More importantly, \textit{TreatStitch w/ SB} exhibited substantial additional gains under this restricted data setting. This is likely because, in the sparse offline dataset, the number of valid stitching points between trajectories is limited. Our SB method addresses this challenge by constructing bridging trajectories that enable stitching even between dissimilar intermediate states, thereby increasing the availability of viable stitching points and further enhancing model performance through additional data augmentation. Moreover, to assess the versatility and generalizability of \textit{TreatStitch}, we integrated it with other offline RL models beyond CQL. The corresponding results are in Appendix~\textcolor{red}{\ref{additional_results}}.

\subsubsection{Real-World Data Results.} 
We also evaluated our framework on the MIMIC-III dataset, which includes real-world clinical data on sepsis treatment in the ICU. We compared our \textit{TreatStitch} against strong competing methods, including GTA, ATraDiff, and RTDiff. As demonstrated in Table~\ref{mimic_table}, our \textit{TreatStitch} framework consistently outperformed all competing methods, even when evaluated using clinically meaningful metrics such as SOFA and NEWS2, alongside standard off-policy evaluation metrics like WIS and DR. These results highlight the practical applicability and robustness of our framework when applied to real-world clinical data beyond benchmark settings. Additionally, \textit{TreatStitch w/ SB} achieved a modest further performance gain, suggesting the potential benefits of SB method in real-world clinical data. 

\begin{figure}[t]
\includegraphics[width=0.95\columnwidth]{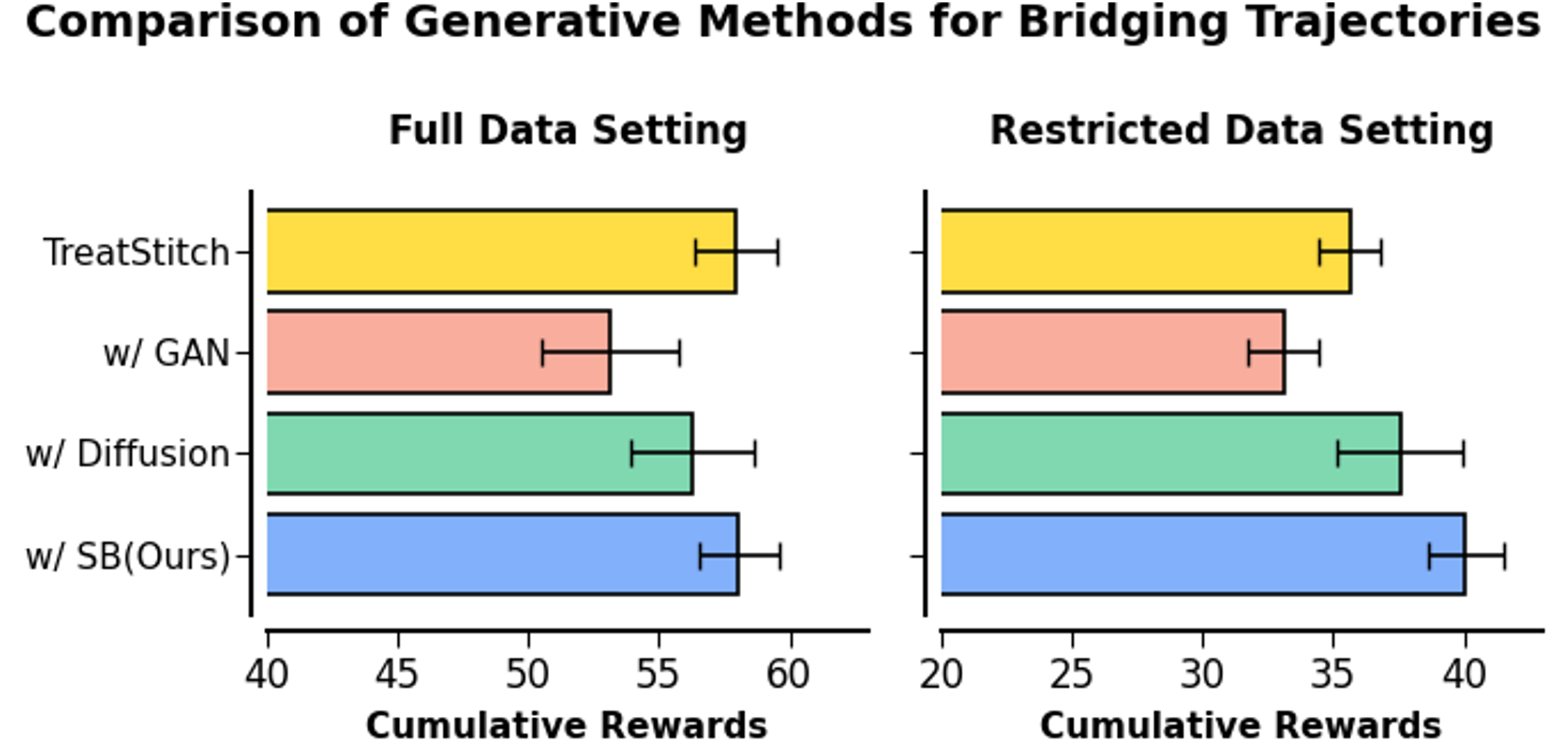}
\centering
\caption{Comparison of various generative methods.}  
\label{discussion_figure}
\end{figure}

\begin{figure}[t]
\includegraphics[width=0.93\columnwidth]{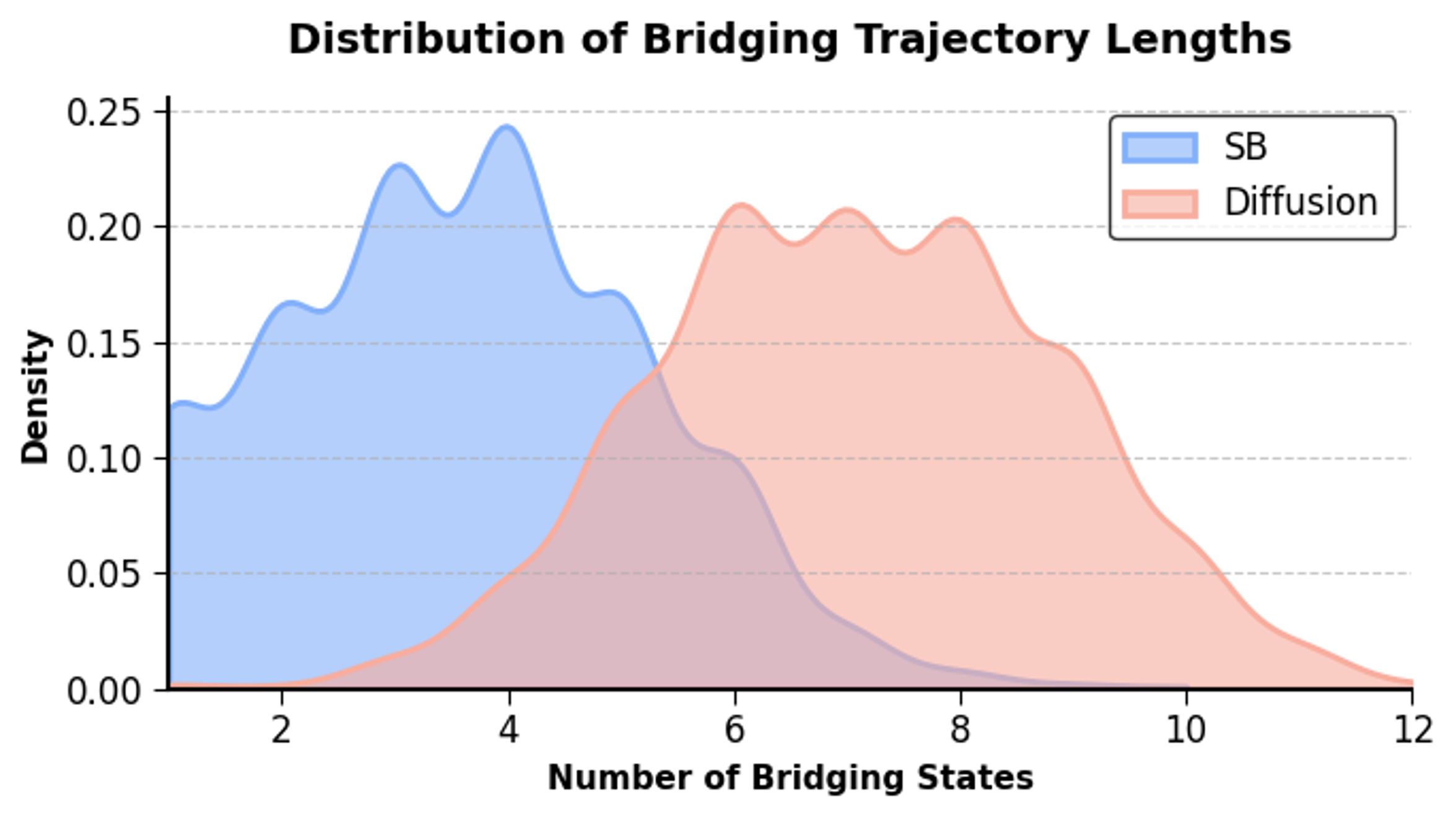}
\centering
\caption{Distribution of bridging trajectory lengths.}  
\label{distribution_figure}
\end{figure} 

\subsubsection{Analysis of Bridging Trajectories.} To analyze the efficacy of our SB method, we explored alternative generative methods for constructing bridging trajectories. Specifically, we developed two additional variants: \textit{TreatStitch w/ GAN} and \textit{TreatStitch w/ Diffusion}, the latter drawing inspiration from prior work \cite{diffstitch}.  As in Figure~\ref{discussion_figure}, the GAN method showed a significant performance drop in both full and restricted data settings, reaffirming the fundamental limitations of GAN in generating high-quality treatment data. While the diffusion method yielded comparable results, our SB method outperformed both methods across all settings. To provide quantitative evidence, we compared the distribution of bridging trajectory lengths generated by SB and diffusion methods. Specifically, we counted the number of generated bridging states for each method and visualized their probability distributions using kernel density estimation. As in Figure~\ref{distribution_figure}, our SB method displayed a pronounced peak at shorter lengths, indicating it generated shorter bridging trajectories than the diffusion method. This can be attributed to the fact that SB is based on OT theory, which aims to identify the smooth and energy-efficient trajectories between states.

\section{Conclusion}
In this work, we propose \textit{TreatStitch}, a novel data augmentation framework designed to enhance offline RL for adaptive treatment strategies (ATS) in clinical settings. Unlike traditional methods that synthesize data from scratch, \textit{TreatStitch} generates clinically valid synthetic trajectories by `stitching' segments of existing treatment data. We empirically validate the efficacy of \textit{TreatStitch} on multiple datasets, and theoretically demonstrate its ability to mitigate OOD transitions.

\section{Acknowledgement}
This work was supported by the Ministry of Science and ICT (MSIT), Korea, through the Institute of Information \& Communications Technology Planning \& Evaluation (IITP) [Artificial Intelligence Graduate School Program at Korea University, No. RS-2019-II190079]; the National Research Foundation of Korea (NRF) [No. RS-2023-00212498]; and the Korea Health Industry Development Institute (KHIDI) under the Federated Learning-based Drug Discovery Acceleration Project (K-MELLODDY) [No. RS-2025-16066488] and the Frailty Zero Project [No. RS-2025-25455839].

\section*{Contribution Statement}
Dong-Hee Shin is the first author, and Tae-Eui Kam is the corresponding author. Deok-Joong Lee contributed to the discussion, and Young-Han Son assisted with proofreading. All authors have read and approved the final manuscript.


\cleardoublepage
\appendix
\setcounter{secnumdepth}{2}
\renewcommand{\thesection}{\Alph{section}}

\makeatletter
\renewcommand{\@seccntformat}[1]{\csname the#1\endcsname.\hspace{0.5em}}
\makeatother

\twocolumn[
\begin{center}
\LARGE \textbf{Appendix (Supplementary Material)}
\end{center}
\vspace{1em}
]

\section{Extended Related Work and Preliminary}
\label{extended_related_work}

\subsection{Adaptive Treatment Strategies}
Adaptive treatment strategies (ATS), also known as dynamic treatment regimes, indicate sequential clinical decision-making frameworks that aim to optimize treatment plans by adapting them over time in response to a patient’s evolving symptoms and medical history. In ATS, treatment decisions are not static but instead depend on previous patient responses, with the ultimate goal of optimizing long-term clinical outcomes rather than short-term symptom relief. As mentioned in the main manuscript, the application of conventional online reinforcement learning (RL) in clinical settings is severely constrained by ethical and safety concerns. The `trial-and-error' learning mechanism inherent to online RL—where the algorithm learns by experimenting with different treatment options directly on patients—is strictly prohibited in clinical settings due to risks of harm to patients. Consequently, offline RL has emerged as the de facto standard method for optimizing ATS. By learning policies exclusively from pre-collected historical treatment data, offline RL circumvents the need for online trial-and-error, making it a safer and more practical approach for clinical settings.

Most prior works in this ATS have focused on developing algorithmic improvements tailored to specific clinical tasks. Specifically, advanced offline RL algorithms have been proposed for treatment recommendation in domains such as anesthesia control \cite{jbhi_ats} and mechanical ventilation \cite{aaai_ats}. These methods generally focus on improving the algorithmic architecture or learning dynamics of offline RL methods, aiming to improve their performance when applied to specialized domains. However, a well-known limitation of RL methods is their substantial demand for large and diverse datasets to train effective policies \cite{rl_book,shin_bci}. This requirement poses a significant challenge in clinical settings, where high-quality longitudinal treatment data is often scarce, expensive to collect, and restricted by privacy regulations. Despite this data bottleneck, data-centric strategies in ATS applications have gained relatively little attention. 

To address this gap, we propose a novel data augmentation framework specifically designed to enhance offline RL in ATS applications. Rather than introducing yet another RL algorithm, our approach focuses on enriching the training dataset by generating clinically valid synthetic treatment trajectories through a technique we call \textit{treatment stitching}. This method constructs new trajectories by recombining real patient treatment segments in a way that preserves clinical validity. Our approach enables clinicians and researchers to apply existing high-performing offline RL models to their specific ATS tasks, while leveraging our framework to mitigate limitations in data availability. By incorporating these clinically valid stitched trajectories into the offline dataset, our framework enriches the training data and facilitates more effective policy learning, thereby improving offline RL performance under constrained treatment data settings.

\subsection{Data Augmentation Methods for Offline RL}
Data augmentation has become a widely adopted strategy in the offline RL domain, where direct interaction with the environment is not permitted. By increasing the diversity of training data, data augmentation helps mitigate overfitting and improves value function approximation, leading to more robust policy learning \cite{offline_rl1}. In recent years, the success of generative models has further driven their use for data augmentation in offline RL. One of the most straightforward approaches involves training generative models, such as GANs \cite{gan} or diffusion models like DDPM \cite{ddpm}, directly on the offline dataset. Then, synthetic data can be generated by simply sampling from the learned distribution of these generative models (e.g., using \lstinline[basicstyle=\ttfamily\small]{model.sample()}). While this approach is simple and intuitive, such methods often lack explicit guidance or regularization, making them prone to generating low-quality or irrelevant data, which can ultimately degrade policy learning \cite{molstitch}.

To address these limitations, more advanced data augmentation methods have been proposed. For instance, SynthER \cite{synther} introduces synthetic experience replay by using a diffusion model to upsample the offline dataset in a flexible manner. However, SynthER tends to focus on individual transitions and may not fully capture the longer-term dependencies at the trajectory level. To tackle this, methods like GTA \cite{gta} and ATraDiff \cite{atradiff} have been introduced. Specifically, GTA leverages diffusion models for data augmentation while using amplified return guidance to produce higher-reward trajectories that enhance the quality of augmented data. ATraDiff, on the other hand, leverages a coarse-to-fine strategy and employs multiple diffusion models to generate diverse samples with varying lengths. More recently, RTDiff \cite{rtdiff} proposes the novel idea of generating synthetic trajectories in reverse order, aiming to reduce overestimation bias and improve learning stability. Hence, RTDiff has achieved state-of-the-art performance in general RL tasks.

\subsection{Data Augmentation in Clinical Settings}
While the aforementioned data augmentation methods have demonstrated promising results in general RL tasks, they are not designed with the unique requirements of clinical settings in mind. In particular, these methods typically rely on generative models to synthesize trajectories entirely from scratch. Although effective in domains such as robotics or games, this strategy may lead to the generation of clinically invalid trajectories when applied to healthcare domains. This is because clinical treatment data is often shaped by complex physiological constraints, long-term causal dependencies, and ethical considerations. To address this gap, we introduce a novel data augmentation framework specifically designed for ATS in clinical settings. By identifying similar intermediate states across different trajectories, our framework `cuts' and `stitches' existing trajectory segments to generate synthetic trajectories that maintain clinical validity.

\clearpage
\section{Results for Backbone Offline RL Methods}
\label{appendix_backbone}
In this study, we selected CQL \cite{cql} as our main backbone offline RL method due to its demonstrated superior performance compared to other methods. To provide quantitative evidence supporting this selection, we evaluated the performance of various offline RL methods on the EpiCare benchmark, presented in Table \ref{backbone_table}. It is important to note that the first three methods—Random Policy, Standard of Care (SoC), and Behavior Cloning (BC)—are not offline RL methods but serve as crucial reference points for performance evaluation. Specifically, the Random Policy selects actions uniformly at random and establishes a lower-bound baseline for performance. The SoC represents the minimum standard of effectiveness expected in clinical practice and is commonly used as a baseline in healthcare research. Surpassing SoC implies a clinically meaningful improvement. The BC is a supervised learning approach that aims to replicate clinicians' historical decisions by minimizing the discrepancy between predicted and actual treatment decisions.

As shown in Table \ref{backbone_table}, actor-critic based offline RL methods, such as EDAC \cite{edac} and AWAC \cite{awac}, underperformed compared to the BC method. This observation suggests that, without careful algorithmic design, RL does not inherently guarantee improvements over supervised learning in clinical settings. TD3+BC \cite{td3}, which integrates offline RL with behavior cloning, did surpass the SoC baseline and achieved comparable results, but its performance remained suboptimal. In contrast, value-based offline RL methods—DQN \cite{offline_dqn}, IQL \cite{iql}, and CQL \cite{cql}—consistently and significantly outperformed the SoC baseline, demonstrating strong overall performance. Among these, CQL achieved the best results, validating our choice of CQL as the primary backbone model for our experiments. While CQL achieved the highest performance, both DQN and IQL also proved to be robust alternatives. To further demonstrate the versatility and generalizability of our \textit{TreatStitch} framework, we conducted additional experiments using DQN and IQL as backbone models; their results are in Appendix \textcolor{red}{\ref{additional_results}}.

\begin{table}[h]
\scriptsize
\centering
\renewcommand{\arraystretch}{1.5} 
\setlength{\tabcolsep}{1.9pt} 
\begin{tabular}{c|cccccccc|c}
\noalign{\smallskip}\noalign{\smallskip} \hline\hline
\textbf{Method}   & \textbf{Env1}   & \textbf{Env2}   & \textbf{Env3}   & \textbf{Env4}   & \textbf{Env5}   & \textbf{Env6}   & \textbf{Env7}   & \textbf{Env8}    & \textbf{Mean}    \\
\cline{1-10}
Random Policy & 2.53 & -8.06 & -19.36 & -0.23 & 7.40 & -1.48 & -4.53 & -7.50 & -3.90 \\
Standard of Care  & 35.26 & 33.95  & 19.46  & 27.83  & 29.21  &  34.55 & 28.17  & 30.82  & 29.90  \\
Behavior Cloning & 34.65  & 20.34  & 15.54  & 14.16  & 23.02  & 1.52  &  9.69 & 6.76  & 15.71 \\
\cline{1-10}
EDAC & 28.46  & 24.15  & -5.01  & 32.25  & 30.27  & -7.81  & -27.23  & -14.46  & 7.58  \\
AWAC & 26.71  & 23.68 & 5.08  & 9.03  & 19.38  & 9.19  & 0.37  & -4.31 & 11.14  \\
TD3+BC & 53.01  & 50.86  & 2.32  & 40.78  & 31.04  & 39.40  & 35.53  & 27.25   & 35.02  \\
DQN & 57.44  & 53.07  & 47.80  & \textbf{58.16}  & 51.64  & 53.58  & 53.33  & 48.26  & 52.91  \\
IQL  & \textbf{62.55}  & \textbf{58.40}  & \textbf{50.03}  & 52.06  & 53.84  & 52.14   & 52.40 & 51.19  & 54.07  \\
CQL & 59.65  & 57.03   & 47.25  & 53.82  & \textbf{54.35}   & \textbf{54.68}  & \textbf{55.59}  & \textbf{52.55}  & \textbf{54.37}  \\
\hline \hline
\end{tabular}
\caption{Comparison of average cumulative rewards for each backbone offline RL method on the EpiCare benchmark.}
\label{backbone_table}
\end{table}

\section{Additional Results}
\label{additional_results}
Note that our proposed \textit{TreatStitch} framework is designed to be model-agnostic and easily adaptable to various offline RL methods in a plug-and-play manner. To demonstrate the versatility and generalizability of \textit{TreatStitch}, we conducted additional experiments using two alternative backbone offline RL methods: IQL and DQN, both of which have shown competitive performance on the EpiCare benchmark. We evaluated each method under both the full data and restricted data settings. The results are presented in Tables~\ref{iql_full} and~\ref{iql_restricted} for IQL, and Tables~\ref{dqn_full} and~\ref{dqn_restricted} for DQN.

For the IQL backbone, we observed that existing augmentation methods such as GTA and ATraDiff provided no or only modest improvements over the IQL backbone. Among the competing methods, RTDiff achieved the strongest performance, but still showed limited gains. In contrast, our \textit{TreatStitch} demonstrated substantial performance improvements across most environments when used in conjunction with IQL, confirming its compatibility and effectiveness as a data augmentation strategy beyond CQL. Moreover, the extended version, \textit{TreatStitch w/ SB}, further achieved additional improvements in performance. Interestingly, similar to our main experiments, the benefits of \textit{TreatStitch w/ SB} were more noticeable in the restricted data setting, where data is scarce. We also observed similar patterns when using DQN as the backbone as shown in Tables~\ref{dqn_full} and~\ref{dqn_restricted}. Our \textit{TreatStitch} consistently improved performance and outperformed other augmentation methods, highlighting its robustness.

\begin{table}[h]
\scriptsize
\centering
\renewcommand{\arraystretch}{1.5} 
\setlength{\tabcolsep}{1.9pt} 
\begin{tabular}{c|cccccccc|c}
\noalign{\smallskip}\noalign{\smallskip} \hline\hline
\textbf{Method}   & \textbf{Env1}   & \textbf{Env2}   & \textbf{Env3}   & \textbf{Env4}   & \textbf{Env5}   & \textbf{Env6}   & \textbf{Env7}   & \textbf{Env8}    & \textbf{Mean}    \\
\cline{1-10}
\multicolumn{1}{l|}{IQL (Backbone)} & 62.55 & 58.40   & 50.03  & 52.06  & 53.84   & 52.14  & 52.40  & 51.19  & 54.07  \\
\multicolumn{1}{l|}{+ GTA} & 61.80  & 50.18 & 46.00 & 52.68 & 57.10 & 51.50 & 56.65 & 50.52 & 53.30 \\
\multicolumn{1}{l|}{+ ATraDiff} & 59.96 & 58.97 & 50.32 & \textbf{53.82} & 55.54 & 53.07 & 52.63 & 51.62 & 54.49  \\
\multicolumn{1}{l|}{+ RTDiff} & 61.63  & 60.07 & 50.41 & 52.63 & 56.38 & 54.37 & 52.38 & 51.56 & 54.93 \\
\cline{1-10}
\multicolumn{1}{l|}{\textbf{+ TreatStitch}} & 63.60  & 62.31 & \textbf{52.75} & 53.48 & 56.91 & 55.46 & 56.95 & 53.39 & 56.85  \\
\multicolumn{1}{l|}{\textbf{+ TreatStitch w/ SB}} & \textbf{64.91} & \textbf{63.10} & 52.71 & 53.72 & \textbf{57.98} & \textbf{56.63} & \textbf{57.33} & \textbf{53.97}  & \textbf{57.54} \\

\hline \hline
\end{tabular}
\caption{Performance comparison of each method using the IQL backbone on EpiCare under the \textbf{full data} setting.}
\label{iql_full}
\end{table}

\begin{table}[h]
\scriptsize
\centering
\renewcommand{\arraystretch}{1.5} 
\setlength{\tabcolsep}{1.9pt} 
\begin{tabular}{c|cccccccc|c}
\noalign{\smallskip}\noalign{\smallskip} \hline\hline
\textbf{Method}   & \textbf{Env1}   & \textbf{Env2}   & \textbf{Env3}   & \textbf{Env4}   & \textbf{Env5}   & \textbf{Env6}   & \textbf{Env7}   & \textbf{Env8}    & \textbf{Mean}    \\
\cline{1-10}
\multicolumn{1}{l|}{IQL (Backbone)} & 20.29  & 18.96 & 8.00  & 12.30 & 20.83 & 15.05 & 8.91 & 14.32 & 14.83 \\
\multicolumn{1}{l|}{+ GTA} & 26.47 & 22.55 & 11.75  & 12.20 & 24.85 & 16.19 & 10.03 & 16.90 & 17.62  \\
\multicolumn{1}{l|}{+ ATraDiff} & 25.27 & 25.23 & 9.82  & 10.45 & 29.49 & 18.76 & 11.33 & 15.62 & 18.25 \\
\multicolumn{1}{l|}{+ RTDiff} & 27.53 & 28.35 & 16.56  & 15.44 & 31.92 & 21.69 & 13.10 & 16.99 & 21.45 \\
\cline{1-10}
\multicolumn{1}{l|}{\textbf{+ TreatStitch}} & 42.41 & 32.47 & 31.65  & 29.55 & 30.29 & 32.68 & 34.85 & 36.63 & 33.82  \\
\multicolumn{1}{l|}{\textbf{+ TreatStitch w/ SB}}& \textbf{45.18} & \textbf{36.93} & \textbf{33.16}  & \textbf{33.93}& \textbf{36.35} & \textbf{35.11} & \textbf{38.45} & \textbf{37.00} & \textbf{37.01} \\

\hline \hline
\end{tabular}
\caption{Performance comparison of each method using the IQL backbone on EpiCare under the \textbf{restricted data} setting.}
\label{iql_restricted}
\end{table}

\begin{table}[h]
\scriptsize
\centering
\renewcommand{\arraystretch}{1.5} 
\setlength{\tabcolsep}{1.9pt} 
\begin{tabular}{c|cccccccc|c}
\noalign{\smallskip}\noalign{\smallskip} \hline\hline
\textbf{Method}   & \textbf{Env1}   & \textbf{Env2}   & \textbf{Env3}   & \textbf{Env4}   & \textbf{Env5}   & \textbf{Env6}   & \textbf{Env7}   & \textbf{Env8}    & \textbf{Mean}    \\
\cline{1-10}
\multicolumn{1}{l|}{DQN (Backbone)} & 57.44  & 53.07  & 47.80  & 58.16  & 51.64  & 53.58  & 53.33  & 48.26  & 52.91  \\
\multicolumn{1}{l|}{+ GTA} & 57.30 & 52.58 & 49.05 & 54.26 & 49.79 & 52.83 & 51.11 & 50.06 & 52.12 \\
\multicolumn{1}{l|}{+ ATraDiff} & 59.02  & 53.05 & 46.50 & 54.57 & 51.53 & 55.26 & 54.54 & 50.13 & 53.08 \\
\multicolumn{1}{l|}{+ RTDiff} & 60.50  & 54.11 & 46.66 & 55.59 & \textbf{53.78} & 56.08 & 56.28 & 51.50 & 54.31 \\
\cline{1-10}
\multicolumn{1}{l|}{\textbf{+ TreatStitch}} &  63.26 & \textbf{59.98} & 52.18 & 58.15 & 52.57 & 56.06 & \textbf{59.29} & 53.18 & 56.83 \\
\multicolumn{1}{l|}{\textbf{+ TreatStitch w/ SB}} & \textbf{64.35} & 59.02 & \textbf{52.71} & \textbf{58.35} & 53.34 & \textbf{57.49} & 59.00 & \textbf{54.74} & \textbf{57.38} \\

\hline \hline
\end{tabular}
\caption{Performance comparison of each method using the DQN backbone on EpiCare under the \textbf{full data} setting.}
\label{dqn_full}
\end{table}

\begin{table}[h]
\scriptsize
\centering
\renewcommand{\arraystretch}{1.5} 
\setlength{\tabcolsep}{1.9pt} 
\begin{tabular}{c|cccccccc|c}
\noalign{\smallskip}\noalign{\smallskip} \hline\hline
\textbf{Method}   & \textbf{Env1}   & \textbf{Env2}   & \textbf{Env3}   & \textbf{Env4}   & \textbf{Env5}   & \textbf{Env6}   & \textbf{Env7}   & \textbf{Env8}    & \textbf{Mean}    \\
\cline{1-10}
\multicolumn{1}{l|}{DQN (Backbone)} & 14.03 & 19.72 & 17.22 & 15.35 & 21.58  & 13.48 & -1.80 & 2.88 & 12.81  \\
\multicolumn{1}{l|}{+ GTA} & 15.20 & 19.28 & 18.95 & 17.73 & 23.11  & 14.92 & 3.01 & 4.05 & 14.53  \\
\multicolumn{1}{l|}{+ ATraDiff} & 22.76 & 24.54 & 18.99 & 19.27 & 20.11  & 19.82 & 3.41 & 7.07 & 17.00  \\
\multicolumn{1}{l|}{+ RTDiff} & 24.03 & 30.80 & 18.76 & 21.74 & 24.33  & 20.90 & 8.23 & 11.81 & 20.08  \\
\cline{1-10}
\multicolumn{1}{l|}{\textbf{+ TreatStitch}} & 42.66 & 35.67 & 32.93 & 28.68 & 30.73  & 36.53  & 32.54 & 28.01 & 33.47  \\
\multicolumn{1}{l|}{\textbf{+ TreatStitch w/ SB}} & \textbf{49.47} & \textbf{36.16} & \textbf{36.92} & \textbf{33.31} & \textbf{33.95}  & \textbf{40.74} & \textbf{38.17} & \textbf{30.00} & \textbf{37.34}  \\

\hline \hline
\end{tabular}
\caption{Performance comparison of each method using the DQN on EpiCare under the \textbf{restricted data} setting.}
\label{dqn_restricted}
\end{table}

\section{Detailed Analysis of Priority Sampling}
\label{appendix_priority_sampling}
In this work, we employed priority sampling to strategically select trajectories, motivated by its demonstrated effectiveness in other domains \cite{dymol}. Here, we present a detailed justification for this design choice and evaluate its effectiveness through empirical analysis.

A straightforward alternative to priority sampling is uniform random sampling, where trajectories are selected at random from the high-reward and low-reward groups without any preference. To evaluate the advantage of our priority-based sampling, we conducted additional experiments comparing the performance of priority sampling versus random sampling within the \textit{TreatStitch} framework.

As shown in Table~\ref{table:priority_sampling}, while random sampling yielded reasonably strong performance, it consistently underperformed compared to the priority sampling strategy. We attribute this performance difference to a concept analogous to prioritized experience replay (PER) \cite{per}. In PER, transitions that are more surprising or lead to larger Bellman errors are sampled more frequently in order to enhance sample efficiency and accelerate learning dynamics, especially in early training phases or sparse-reward settings.

In the context of our main ATS experiments, a uniform random sampling strategy would treat all treatment trajectories equally, sampling uniformly from both the high-reward and low-reward groups without considering the relative importance or learning potential of different trajectory combinations. This approach may result in many redundant trajectory pairs that provide limited learning value to the offline RL agent. In contrast, priority sampling enables the selection of higher-reward trajectories from $\mathcal{D}_{\text{high}}$ and lower-reward trajectories from $\mathcal{D}_{\text{low}}$ during early training stages. This strategy creates more informative stitched trajectories by combining segments from initially adverse treatment paths with beneficial outcomes, providing richer learning experiences for the agent. Furthermore, our priority sampling strategy incorporates a temperature parameter $\alpha$. By gradually increasing $\alpha$ as training progresses, we effectively shift the sampling strategy from a highly prioritized approach towards more uniform sampling. This approach allows the agent to initially learn from the most informative trajectory combinations and then progressively explore more comprehensive coverage of the offline dataset by the later stages.

\begin{table}[t]
\scriptsize
\centering
\renewcommand{\arraystretch}{1.5} 
\setlength{\tabcolsep}{1.9pt} 
\begin{tabular}{c|cccccccc|c}
\noalign{\smallskip}\noalign{\smallskip} \hline\hline
\textbf{Method}   & \textbf{Env1}   & \textbf{Env2}   & \textbf{Env3}   & \textbf{Env4}   & \textbf{Env5}   & \textbf{Env6}   & \textbf{Env7}   & \textbf{Env8}    & \textbf{Mean}    \\
\cline{1-10}
\multicolumn{1}{l|}{Priority Sampling} & \textbf{65.15} & 62.05  & \textbf{50.57}  & 58.22  & \textbf{57.88}  & \textbf{56.98}  & \textbf{59.09}  & \textbf{53.51}  & \textbf{57.93}  \\
\multicolumn{1}{l|}{Random Sampling} & 65.06  & \textbf{63.01}   & 49.50 & \textbf{58.60} & 57.26 & 55.70 & 58.23 & 52.00 & 57.42  \\
\hline \hline
\end{tabular}
\caption{Performance comparison between priority sampling and random sampling within our \textit{TreatStitch} on EpiCare.}
\label{table:priority_sampling}
\end{table}

Nonetheless, we acknowledge that priority sampling introduces additional implementation complexity. If implementation complexity is a significant concern for the user, a simpler random sampling remains a strong and viable alternative that still provides reasonable performance, as demonstrated in our experiments. The choice between priority sampling and random sampling can be viewed as a trade-off between implementation simplicity and performance optimization. Users seeking a simpler implementation can adopt random sampling and still benefit from the core principles of our stitching-based data augmentation framework.


\section{Detailed Explanation of the Schrödinger Bridge for Treatment Stitching}
\label{extended_SB}

\subsection{Schrödinger Bridge problem}

The Schrödinger Bridge (SB) problem is the classical problem of finding an optimal stochastic process that connects two given probability distributions \cite{shi2023diffusion}. The SB problem aims to find the path measure of a stochastic process that connects the start and target distributions while remaining closest to a reference stochastic process in terms of Kullback–Leibler (KL) divergence. In the generative modeling context, the start distribution is typically matched at $t=0$, and the target at $t=T$. Therefore, the SB problem can be mathematically formulated as follows:
\begin{equation*}
\mathbb{P}^* = \underset{\mathbb{P}}{\arg\min} \, \mathrm{KL}(\mathbb{P} \| \mathbb{Q}) \; \text{s.t.} \; \mathbb{P}_0 = p_{\text{start}}, \mathbb{P}_T = p_{\text{target}}.
\end{equation*}
Note that the SB problem can be viewed as a formulation of entropy-regularized optimal transport \cite{shi2023diffusion}. In particular, under an appropriate choice of cost function (e.g., quadratic cost) and reference stochastic process (e.g., Brownian motion), the SB problem solution coincides with the optimal solution of the entropy-regularized OT problem \cite{de2021diffusion}.

\subsection{Reformulation of the SB problem via Stochastic Differential Equations}
Since the path measure is often difficult to handle directly, the SB problem can be reformulated using stochastic differential equations (SDEs) \cite{chen2021likelihood}. Following \cite{chen2021likelihood}, the SB problem characterizes a stochastic optimal control programming with energy (e.g., $(\tfrac{1}{2}\lVert \mathbf{u} \rVert)^2$), minimization, where the optimization is given as follows:
\begin{equation}
\arraycolsep=1.4pt
\begin{array}{c}
\displaystyle \min_{\mathbf{u}}\, \mathbb{E} \left[ \int_0^T \tfrac{1}{2} \left\lVert \mathbf{u}(\mathbf{X}_t, t) \right\rVert^2 dt \right] \\
\text{s.t.} 
\begin{cases}
\mathrm{d}\mathbf{X}_t = \left[ f(\mathbf{X}_t, t) + g(t)\mathbf{u}(\mathbf{X}_t, t) \right] \mathrm{d}t +g(t)\mathrm{d}\mathbf{W}_t \\
\mathbf{X}_0 \sim p_0(\mathbf{X}),\quad \mathbf{X}_T \sim p_T(\mathbf{X})
\end{cases},
\end{array}
\label{sb_control}
\end{equation}
where $g$ is uniformly lower-bounded and $f$ satisfies Lipschitz conditions \cite{chen2021likelihood}. Then, the Hopf-Cole transformation applied to Equation \ref{sb_control} results in Equation \ref{schrodinger_system}, which characterizes the optimal control solution as a coupled Schrödinger system consisting of a pair of partial differential equations \cite{caluya}:
\begin{equation*}
\begin{cases}
\frac{\partial \Psi}{\partial t} = -\langle \nabla \Psi, f \rangle - \frac{1}{2} g^2 \Delta \Psi \\
\frac{\partial \hat{\Psi}}{\partial t} = -\nabla \cdot [\hat{\Psi} f] + \frac{1}{2} g^2 \Delta \hat{\Psi}
\end{cases}
\hspace{-3mm} \text{s.t.} \hspace{-0.5mm}
\begin{cases}
\Psi(\cdot, 0)\hat{\Psi}(\cdot, 0)\!=\!p_{\text{start}}, \\
\Psi(\cdot, T)\hat{\Psi}(\cdot, T)\!=\!p_{\text{target}}.
\end{cases}
\end{equation*}
In Equation \ref{schrodinger_system}, $\Psi$ and $\hat{\Psi}$ denote the Schrödinger potentials, which are solutions to the Schrödinger system. The optimal control solution $\mathbf{u}^*$ is given by
\begin{equation}
\begin{aligned}
\mathbf{u}^* &= g\ \nabla \log\Psi.
\end{aligned}
\label{optimal_control}
\end{equation}
By substituting the optimal control $\mathbf{u}^*$ with the optimal potentials $\Psi$ and $\hat{\Psi}$ in Equation \ref{sb_control}, the SB problem solution can be expressed as a pair of forward–backward stochastic differential equations\footnote{For brevity, we omit input variables $ \mathbf{u}\equiv\mathbf{u}(\mathbf{X}_t, t), f\equiv f(\mathbf{X_t, t}),  g \equiv g(t), \Psi \equiv \Psi(\mathbf{X}_t, t), \hat{\Psi} \equiv \hat{\Psi}(\mathbf{X}_t, t)$.} \cite{chen2021likelihood}:
{\small
\begin{equation}
\begin{aligned}
&\text{Forward SDE:} \\
&\quad \mathrm{d}\mathbf{X}_t = \left[ f + g^2 \nabla \log \Psi(\mathbf{X}_t, t) \right] \mathrm{d}t + g\,\mathrm{d}\mathbf{W}_t, \quad \mathbf{X}_0 \sim p_{\text{start}}, \\
&\text{Backward SDE:} \\
&\quad \mathrm{d}\mathbf{X}_t = \left[ f - g^2 \nabla \log \hat{\Psi}(\mathbf{X}_t, t) \right] \mathrm{d}t + g\,\mathrm{d}\mathbf{W}_t, \quad \mathbf{X}_T \sim p_{\text{target}},
\end{aligned}
\end{equation}
}
where $\nabla \log \Psi(\mathbf{X}_t, t)$ and $\nabla \log \hat{\Psi}(\mathbf{X}_t, t)$ correspond to the optimal drifts in the forward and backward processes of SB \cite{chen2021likelihood}. These two equations are related via a reverse-time formulation, and have been widely applied in distribution matching tasks such as dataset interpolation and image translation \cite{de2021diffusion, theodoropoulos2024feedback}.

\section{Training Objective of the Schrödinger Bridge for Bridging State Generation}

\label{appendix_obj}

This section provides the detailed training objective and procedure for the neural network $G_\phi$, which is trained to generate a bridging state $s_{\text{bridge}} = \{\tilde{s}_1, \tilde{s}_2, \dots, \tilde{s}_K \}$. The goal is to generate a smooth transition from the start state $s_t^C$ to the target state $s_{t'}^D$, facilitating the stitching between $\tau_C$ and $\tau_D$.

Note that throughout this section, we use $\mathbf{x}$ and $s$ interchangeably to denote states, where $\mathbf{x}$ represents the general mathematical formulation and $s$ refers to specific states in the context of trajectories (e.g., $s_t^C$, $s_{t'}^D$, $\tilde{s}_k$). Both notations refer to the same mathematical objects in the state space.

To generate the bridging state that enables a smooth transition between $s_t^C$ and $s_{t'}^D$, the network $G_\phi$ is trained to solve the Schrödinger system that captures the underlying stochastic trajectory based on the optimal transport theory.

\subsection{Schrödinger Bridge versus Diffusion Models}
First, we want to clarify the difference between Schrödinger Bridge and typical diffusion models. Although both can be described by SDE equations, they exhibit a fundamental distinction in their objectives, which has significant implications for generating efficient state transitions.

A standard diffusion model learns the time-reversed SDE of a predefined noising process that gradually transforms a complex data distribution into a simple prior, such as a Gaussian. The resulting trajectories are obtained by reversing the noising process. However, they are not explicitly optimized for efficiency, often leading to unnecessarily long or indirect paths through the state space.

In contrast, the Schrödinger Bridge framework directly addresses an \emph{optimal transport} problem. It is uniquely constrained by both a start distribution \( p_{\text{start}} \) and a target distribution \( p_{\text{target}} \). The objective is to identify the stochastic process \( \mathbb{P}^* \) that transports mass between these distributions while minimizing the Kullback–Leibler (KL) divergence with respect to a reference process \( \mathbb{Q} \) (e.g., Brownian motion):
\begin{equation*}
\mathbb{P}^* = \underset{\mathbb{P}}{\arg\min} \, \mathrm{KL}(\mathbb{P} \| \mathbb{Q}).
\end{equation*}
This KL minimization encourages the learned process \( \mathbb{P}^* \) to stay as close as possible to the simpler dynamics of the reference process \( \mathbb{Q} \). As a result, it implicitly regularizes the geometry of the generated trajectories, naturally favoring more direct and efficient stochastic trajectories.

\subsection{Learning the Schrödinger Potentials via Conditional Score Matching}

To solve the Schrödinger system defined in Equation~\ref{schrodinger_system} of the main manuscript, the network $G_\phi$ must learn to approximate the Schrödinger potentials, $\Psi(\mathbf{x}, t)$ and $\hat{\Psi}(\mathbf{x}, t)$. Rather than estimating these potentials directly, our approach is to learn their gradients, which correspond to the score functions $\nabla \log \Psi(\mathbf{x}, t)$ and $\nabla \log \hat{\Psi}(\mathbf{x}, t)$. This is achieved through conditional denoising score matching, which is tailored to our goal of generating smooth bridge states between specific start and target states.

The training set for \( G_\phi \) is constructed from pairs of states $(\mathbf{x}_{\text{start}}, \mathbf{x}_{\text{target}})$ sampled from the trajectories in $\mathcal{D}$. Again, each $\mathbf{x}$ represents a state from some trajectory, equivalent to the $s$ notation used elsewhere. During each training step, a mini-batch of $B$ such pairs is sampled. For each pair \(i\) in the batch, a time \(t_i \sim \mathcal{U}(0, 1)\) is sampled, and the start state is perturbed with Brownian noise scaled by \(\sigma\):
\begin{equation}
\tilde{\mathbf{x}}^{\text{fwd}}_i
= \mathbf{x}_{\text{start},i} + \sigma \sqrt{t_i}\, \boldsymbol{\epsilon}^{\text{fwd}}_i,
\quad \boldsymbol{\epsilon}^{\text{fwd}}_i \sim \mathcal{N}(\mathbf{0}, \mathbf{I}).
\end{equation}
The corresponding target score for the denoising objective is the negative normalized noise:
\begin{equation}
\mathbf{x}^{\star,\text{fwd}}_i
= -\frac{\boldsymbol{\epsilon}^{\text{fwd}}_i}{\sigma \sqrt{t_i} + \epsilon_{\text{stab}}},
\end{equation}
where $\epsilon_{\text{stab}}$ is a small constant for numerical stability.

For symmetry, the target state is similarly perturbed at the complementary time \(1 - t_i\):
\begin{equation}
\tilde{\mathbf{x}}^{\text{bwd}}_i
= \mathbf{x}_{\text{target},i} + \sigma \sqrt{1 - t_i}\, \boldsymbol{\epsilon}^{\text{bwd}}_i,
\quad \boldsymbol{\epsilon}^{\text{bwd}}_i \sim \mathcal{N}(\mathbf{0}, \mathbf{I}),
\end{equation}
with its associated score target defined as:
\begin{equation}
\mathbf{x}^{\star,\text{bwd}}_i
= -\frac{\boldsymbol{\epsilon}^{\text{bwd}}_i}{\sigma \sqrt{1 - t_i} + \epsilon_{\text{stab}}}.
\end{equation}

The model \( G_\phi \) is trained to predict conditional target scores by minimizing both forward and backward denoising objectives. The forward score matching loss trains \( G_\phi \) to approximate the forward score function \( \nabla_{\mathbf{x}} \log \Psi(\mathbf{x}, t) \):
{\small
\begin{equation}
\mathcal{L}_{\text{fwd}}(\phi)
= \frac{1}{B} \sum_{i=1}^{B}
\left\| G_\phi^{\text{fwd}}(\tilde{\mathbf{x}}^{\text{fwd}}_i, t_i \mid \mathbf{x}_{\text{start},i}, \mathbf{x}_{\text{target},i})
- \mathbf{x}^{\star,\text{fwd}}_i \right\|_2^2.
\label{fwd_loss}
\end{equation}
}
Similarly, the backward score matching loss estimates the time-reversed score function \( \nabla_{\mathbf{x}} \log \hat{\Psi}(\mathbf{x}, t) \) by evaluating the model on reversed time inputs:
{\small
\begin{equation}
\begin{aligned}
\mathcal{L}_{\text{bwd}}(\phi)
&= \frac{1}{B}\sum_{i=1}^{B}
\Bigl\lVert G_{\phi}^{\text{bwd}}\bigl(\tilde{\mathbf{x}}^{\text{bwd}}_{i},\, 1-t_i \,\big|\, \mathbf{x}_{\text{start},i},\, \mathbf{x}_{\text{target},i}\bigr) -\, \mathbf{x}^{\star,\text{bwd}}_{i} \Bigr\rVert_{2}^{2}.
\end{aligned}
\label{bwd_loss}
\end{equation}
}
\subsection{Complete Training Objective and Bridging State Generation}
The complete training objective of \( G_\phi \) is the sum of the forward and backward score‑matching losses:
\begin{equation}
\mathcal{L}_{\text{total}}(\phi) = \mathcal{L}_{\text{forward}}(\phi) + \mathcal{L}_{\text{backward}}(\phi).
\end{equation}

Once \( G_\phi \) is trained, a bridging trajectory between states \( s_t^C \) and \( s_{t'}^D \) is generated by simulating a single forward SDE from \( t = 0 \) to \( t = 1 \). The drift of this SDE is corrected at each step by the learned conditional score function.

The generation procedure begins by initializing the state \( \tilde{s}_0 = s_t^C \). Subsequently, for each of the \( K \) discrete time steps, the next state in the sequence is computed using the Euler–Maruyama update rule \cite{euler} for the forward SDE:
\begin{equation}
\tilde{s}_{k+1} = \tilde{s}_k 
+ \tfrac{\sigma^2}{2} \, G_\phi^{\text{fwd}}\!\left(\tilde{s}_k, k \cdot \Delta t \mid s_t^C, s_{t'}^D\right) \cdot \Delta t
+ \sigma \sqrt{\Delta t} \, \boldsymbol{\epsilon}_k,
\end{equation}
where \( k \in \{1, \dots, K\} \), \( \Delta t = 1/K \), \( \sigma \) is the diffusion coefficient, and \( \boldsymbol{\epsilon}_k \sim \mathcal{N}(\mathbf{0}, \mathbf{I}) \) is a standard Gaussian noise vector.
\subsection{Details about Bridging Trajectory Completion}
We acknowledge that our primary focus was on bridging state generation using the SB method. This emphasis arises from the inherent complexity of generating states, which poses greater challenges than inferring actions or rewards. States are typically high-dimensional, time-dependent, and closely tied to the environment’s transition dynamics, which requires maintaining consistency across multiple time steps \cite{rl_book}. In contrast, actions are often low-dimensional (e.g., categorical or discrete), and rewards are usually scalar values that can be normalized within a range.

However, constructing coherent and usable bridging trajectories also requires inferring the appropriate actions and rewards. In the following, we outline the approach used to generate these components based on the bridging states. As outlined in the main manuscript, once the sequence of bridging states $s_{\text{bridge}} = \{\tilde{s}_1, \tilde{s}_2, \dots, \tilde{s}_K\}$ is generated via the SB method, we complete the bridging trajectory by inferring the corresponding actions and rewards. 

To achieve this, we use two dedicated neural networks: an inverse dynamics model $I_\psi(a_t \mid s_t, s_{t+1})$ and a reward prediction model $R_\rho(r_t \mid s_t, a_t)$. These models are trained concurrently using a supervised learning objective over the entire original offline dataset $\mathcal{D}$. The combined loss function guides their learning process such as follows:
{\small
\begin{equation*}
\mathcal{L}(\psi, \rho) = \mathbb{E}_{{\scriptscriptstyle \tau \sim \mathcal{D}}} \left[ 
\left\| I_\psi(s_t, s_{t+1}) - a_t \right\|^2 + 
\left\| R_\rho(s_t, a_t) - r_t \right\|^2 
\right].
\end{equation*}}
For the inverse dynamics model, the objective is to predict the action \( a_t \) that was taken to transition from state \( s_t \) to \( s_{t+1} \). If the action space is discrete, as in our case, this can be framed as a relatively straightforward classification task. Similarly, if the action space is continuous, it is still a simple regression task. The reward prediction model learns to estimate the reward value \( r_t \) associated with a given state-action pair \( (s_t, a_t) \). This is typically framed as a straightforward regression task, especially given that reward values are often normalized, thereby simplifying the learning process.

Once these models \( I_\psi \) and \( R_\rho \) are trained, we apply them to each adjacent pair of bridge states generated by the SB method—\( (\tilde{s}_i, \tilde{s}_{i+1}) \)—to infer the corresponding action \( \tilde{a}_i \) and reward \( \tilde{r}_i \). This process systematically constructs the full bridging trajectory:
\begin{equation*}
\tau_{\text{bridge}} = \left\langle (\tilde{s}_1, \tilde{a}_1, \tilde{r}_1), (\tilde{s}_2, \tilde{a}_2, \tilde{r}_2), \dots, (\tilde{s}_K, \tilde{a}_K, \tilde{r}_K) \right\rangle.
\end{equation*}
Finally, this complete \( \tau_{\text{bridge}} \) is seamlessly integrated with segments from existing real trajectories (\( \tau_C \) and \( \tau_D \)) to construct the additional stitched trajectory \( \tau_{\text{stit}}^{\text{SB}} \), as mentioned in the main manuscript.

\section{Detailed Experimental Settings}
\label{appendix_setting}
In this section, we provide additional details on the experimental setup, including the hyperparameters and the Markov Decision Process (MDP) formulations used for the EpiCare benchmark and the MIMIC-III dataset.

\begin{table}[t]
    \centering
    \footnotesize
    \setlength{\tabcolsep}{6pt}

    \begin{tabular}{lccc}
        \toprule
        \textbf{Hyperparameter} & \textbf{CQL} & \textbf{IQL} & \textbf{DQN} \\
        \midrule
        Batch size & 256 & 256 & 256 \\
        Learning rate & 3e-5 & 3e-4 & 1e-4 \\
        Training steps & 2e5 & 5e5 & 2e5 \\
        Gamma & 0.0 & 1.0 & 0.1 \\
        Dropout & 0.0 & 0.1 & 0.0 \\
        Alpha & 1.0 & - & - \\   
        Beta & - & 3.0 & - \\
        Tau & - & 0.9 & - \\

        \bottomrule
    \end{tabular}
    \caption{Hyperparameter settings for offline reinforcement learning backbone models on the EpiCare benchmark.}
    \label{table:hyperparameters}
\end{table}
\subsection{Hyperparameter Settings}

For our main experiments on the EpiCare benchmark, we closely followed all experimental settings provided in the benchmark without modification. Specifically, we used the hyperparameters provided in the EpiCare benchmark for our main CQL backbone along with IQL and DQN backbone models. The hyperparameters for these offline RL backbone models are provided in Table \ref{table:hyperparameters}. For our additional experiments on the MIMIC-III dataset, we also closely followed all experimental settings provided in a prior work \cite{position_icml}. In these experiments, we also utilized CQL as the main backbone model with the following hyperparameters: a learning rate of 1e-3, a batch size of 256, a stacking number of 3, gamma of 0.99, and an alpha value of 1.0. 

For the treatment stitching process, since the stitching operation is performed by our proposed mechanism rather than a neural network model, no hyperparameters are required for the stitching operation itself. However, there is one crucial hyperparameter that controls the treatment stitching mechanism: the similarity threshold $\delta$. This threshold determines valid stitching points between trajectory segments. If the $\delta$ value is too low, it may compromise clinical validity by allowing stitching between dissimilar intermediate states. Conversely, if $\delta$ is too high, there will be insufficient valid stitching points, preventing the treatment stitching operation from functioning effectively. 

In our study, we set the similarity threshold to $\delta = 0.95$ based on empirical observations. We believe this value was found to be sufficiently high to ensure clinical validity while maintaining adequate stitching opportunities. Across various environments, this threshold value performed consistently well in our experiments. We also observed that values of $\delta > 0.9$ produced comparable results, suggesting that our framework is not overly sensitive to this parameter. Nevertheless, we recommend using $\delta = 0.95$ as it provides a good balance between clinical validity and stitching feasibility.

\subsection{Markov Decision Process (MDP) Formulation}

\subsubsection{EpiCare benchmark.} The MDP formulation for the EpiCare benchmark is defined as follows:

\textbf{State:} The state represents the patient's comprehensive clinical condition at each time step, including disease progression status, adverse events, and clinical symptoms. This encompasses vital signs, disease status indicators, and relevant medical conditions that inform treatment decisions.

\begin{table}[t]
    \centering
    \footnotesize
    \setlength{\tabcolsep}{6pt}
    \begin{tabular}{lcc}
        \toprule
        \textbf{Environment Parameter} & \textbf{Type} & \textbf{Value} \\
        \midrule
        Number of treatments & Integer Value & 16 \\
        Number of disease states & Integer Value & 16 \\
        Number of symptoms & Integer Value & 8 \\
        Remission reward & Continuous Value & 64 \\
        Adverse event penalty & Continuous Value & $-64$ \\
        Adverse event threshold & Continuous Value & 0.999 \\        
        Symptom mean range & Continuous Range & $[0, 2]$ \\
        Symptom std. range & Continuous Range & $[1, 2]$ \\
        Remission probability range & Continuous Range & $[0.8, 1.0]$ \\
        Transition probability range & Continuous Range & $[0.01, 0.2]$ \\
        \bottomrule
    \end{tabular}
    \caption{Key parameter configurations used in EpiCare.}
    \label{table:env_params}
\end{table}
\textbf{Action:} The action space represents the available treatment decisions that clinicians can prescribe. These actions mimic various therapeutic interventions, which might include different types of injections, specific medications, or other therapeutic procedures. In this EpiCare benchmark setting, a total of 16 distinct treatment actions are available.

\textbf{Reward:} The reward function captures patient responses and clinical outcomes to reflect the effectiveness of treatment decisions. In the EpiCare benchmark, the reward structure is designed as follows: $+64$ for achieving remission (positive clinical outcome), $-64$ for adverse events (negative clinical outcome), and a penalty in the range of $[-1, -4]$ to reflect treatment costs and encourage efficient resource utilization.
This reward design encourages the agent to optimize for positive patient outcomes while penalizing excessive or costly therapeutic interventions. Key parameter settings used in this benchmark are summarized in Table~\ref{table:env_params}.

\subsubsection{MIMIC-III dataset.} The real-world MIMIC-III dataset contains clinical treatment data focused on sepsis management in the ICU. Due to privacy restrictions, the dataset is not publicly distributable and is only available upon request; users must handle it with appropriate care. For experiments using this dataset, the MDP formulation is structured differently to reflect the nature of real-world clinical data.

\textbf{State:} The state representation encompasses comprehensive patient physiological parameters and clinical records captured at 4-hour intervals. The state vector incorporates both continuous measurements such as blood pressure, glucose levels, and body temperature, as well as discrete variables including patient demographics and readmission indicators. The state space is constructed from four primary categories: patient demographic and static characteristics, laboratory test results, physiological vital signs, and fluid intake/output measurements. This multidimensional representation provides a holistic view of the patient's clinical status at each decision point.

\textbf{Action:} The action space is defined over two commonly administered interventions in intensive care: vasopressors and intravenous (IV) fluids. Each intervention is discretized into five dosage levels, corresponding to integer bins from 0 (no treatment) to 4 (increasing dosage). Therefore, an action at each time step is defined as a tuple representing the selected dosage level of both vasopressors and IV fluids. The exact dosage mappings for each bin are detailed in Table~\ref{table:action_ranges}.

\begin{table}[t]
    \centering
    \footnotesize
    \setlength{\tabcolsep}{8pt}

    \begin{tabular}{ccc}
        \toprule
        \textbf{Action} & \textbf{IV range (ml/4h)} & \textbf{VA range (mcg/kg/min)} \\
        \midrule
        1 & 0 & 0 \\
        2 & (0, 50] & (0, 0.08] \\
        3 & (50, 180] & (0.08, 0.22] \\
        4 & (180, 530] & (0.22, 0.45] \\
        5 & (530, $\infty$) & (0.45, $\infty$) \\
        \bottomrule
    \end{tabular}
    \caption{Action ranges for IV fluids and vasopressors (VA), with IV fluids measured in ml/4h and VA in mcg/kg/min.}
    \label{table:action_ranges}
\end{table}

\textbf{Reward:} The reward function integrates multiple clinical metrics to assess patient outcomes. The primary formulation, referred to as the SOFA-based reward, combines the Sequential Organ Failure Assessment (SOFA) score \cite{sofa} and lactate levels to jointly capture the progression of organ dysfunction and metabolic instability. The SOFA-based reward is defined as:
\begin{equation}
\begin{split}
r_t^i =\; & c_0 \left( \mathds{1}_{\kappa_t^i = \kappa_{t+1}^i} \cdot \mathds{1}_{\kappa_{t+1}^i > 0} \right) 
+ c_1 (\kappa_{t+1}^i - \kappa_t^i) \\
& + c_2 \tanh(v_{t+1}^i - v_t^i) 
+ \mathds{1}_{t = T_i} \, r_{\text{outcome}}
\end{split}
\end{equation}
where $\kappa$ represents the SOFA score, $v$ denotes lactate concentration, and the constants $c_0$, $c_1$, $c_2$ are set to $-0.025$, $-0.125$, and $-2$, respectively. The outcome reward $r_{\text{outcome}}$ assigns $+15$ for patient survival and $-15$ for mortality.

As an alternative formulation, a second reward function is based on the National Early Warning Score 2 (NEWS2) \cite{news}, which provides a standardized approach to clinical deterioration assessment. The NEWS2-based reward is formulated as:
\begin{equation}
r_t^i = -r_{\text{NEWS2}} + \mathds{1}_{t=T_i} r_{\text{outcome}}.
\end{equation}
The NEWS2 score is normalized to the interval $[0,1]$ to represent mortality probability, with $r_{\text{outcome}}$ set to $-1$ for patient death and $0$ for survival. 

In addition to clinical-based evaluation metrics, we also employ standard off-policy evaluation (OPE) metrics, including weighted importance sampling (WIS) and doubly robust (DR) estimators. These OPE metrics are particularly valuable in clinical settings where direct online policy deployment is infeasible due to ethical and safety constraints. They provide a means of estimating how a learned policy would perform using only logged historical data collected under a different behavior policy. 

In particular, WIS estimator reweights the observed returns by computing the importance ratio between the target policy and the behavior policy. Intuitively, it adjusts the contribution of each trajectory based on how likely the target policy would have taken the same sequence of actions, giving greater weight to more policy-aligned trajectories. Following prior work \cite{position_icml}, we adopt the trajectory-wise WIS formulation as follows: 
\begin{equation}
    \text{WIS} = \frac{\sum_{i} w^i_{1:T_i} G^i}{\sum_{i} w^i_{1:T_i}}, \quad \text{where } w^i_{1:T_i} = \prod_{t=1}^{T_i} \frac{\pi_{\text{target}}(a_t^{i} \mid s_t^{i})}{\pi_{\text{behavior}}(a_t^{i} \mid s_t^{i})}.
\end{equation}
Here, \( G^i = \sum_{t=1}^{T_i} \gamma^{t-1} r_t^{i} \) denotes the discounted return of trajectory \( i \), and \( w^i_{1:T_i} \) represents its corresponding importance weight.

In contrast, the DR estimator enhances robustness in off-policy evaluation by combining model-based value function estimation with importance sampling corrections. Specifically, it uses a learned Q-function \( \hat{Q}(s, a) \), trained (e.g., using temporal-difference methods such as SARSA) on historical data, to estimate expected returns. The DR estimate for each sample is computed as:
\begin{equation}
\begin{split}
\text{DR}_i = \hat{Q}(s_i, \pi(s_i)) + \frac{\pi(a_i \mid s_i)}{\pi_b(a_i \mid s_i)} \big( r_i + \gamma \hat{Q}(s'_i, \pi(s'_i)) \\
- \hat{Q}(s_i, a_i) \big)
\end{split}
\end{equation}
where \( \pi \) is the target policy, \( \pi_b \) is the behavior policy, and \( \hat{Q} \) is the learned Q-function. The Q-function can be trained using a SARSA-style temporal difference (TD) loss:
\begin{equation}
\mathcal{L}_{\text{TD}} = \text{MSE} \left( r + \gamma Q(s', a'), Q(s, a) \right),
\end{equation}
where MSE denotes the mean squared error. In practice, the model with the lowest TD error on the validation set is selected for evaluation. 
\subsection{Computational Resources}
All experiments were implemented in PyTorch \cite{pytorch} and conducted on an NVIDIA GeForce RTX 3090 GPU with 24 GB memory. Since our approach operates on trajectory data instead of high-dimensional inputs like images or videos, its memory requirements are low, enabling all experiments to be run on a single GPU setup. Moreover, our main trajectory stitching mechanism is computationally lightweight, as it does not require training large neural networks to perform the stitching operation. The extended version, \textit{TreatStitch w/ SB}, incorporates Schrödinger Bridge training; however, since it operates on simple trajectory data rather than generating high-dimensional outputs like images, it remains efficient and fully executable on a single GPU. Overall, the computational overhead is minimal and does not present a practical concern.

\section{Descriptions of the Stitching Mechanisms}
\subsection{Clarifications on the Direct Stitching Process}
In this section, we provide detailed clarifications on our proposed stitching process, which constitutes the core mechanism of our TreatStitch framework. As illustrated in Figure~\ref{fig2}(b) of the main manuscript, the stitching operation is visualized as the merging of two similar states at the stitching point. Specifically, at the stitching location, we identify two highly similar states—$s_t^A$ from Trajectory~A and $s_{t'}^B$ from Trajectory~B—such that $\text{Sim}(s_t^A, s_{t'}^B) \geq \delta$.  

While Eq.~\ref{stitching_equation} presents the stitched trajectory as a sequential concatenation of the two segments for notational simplicity, it is important to note that the high similarity between $s_t^A$ and $s_{t'}^B$ ensures that these two states represent essentially the similar underlying clinical condition. Thus, the transition from $s_{t'}^B$ to $s_{t+1}^A$ is not an abrupt or disjoint jump but a continuous transition within the support of the underlying state distribution. In other words, we treat $s_t^A$ and $s_{t'}^B$ as the stitching point, interpreting them as highly similar states. This allows the transition from $s_{t'}^{B}$ to $s_{t+1}^{A}$ to be considered valid and representative of a natural connection between the two trajectory segments.

\subsection{Clarifications on the SB Stitching}
For the extended stitching mechanism based on Schrödinger bridges (SB), a similar principle applies, as shown in Eq.~\ref{sb_stitching_eq}. The bridging trajectory, denoted as $\tau_{\text{bridge}}$, comprises $K$ generated states $\{\tilde{s}_1, \ldots, \tilde{s}_K\}$ that form a smooth transition between $s_t^C$ and $s_{t'}^D$. The complete stitched trajectory, $\tau_{\text{stit}}^{\text{SB}}$, therefore includes both these bridging states and their associated connecting transitions. Specifically, the transition starts from $s_t^C$ to the entry state $\tilde{s}_1$, proceeds sequentially through each pair of consecutive bridging states $(\tilde{s}_i, \tilde{s}_{i+1})$ for $i = 1, \ldots, K-1$, and finally links the last bridging state $\tilde{s}_K$ to the target state $s_{t'}^D$. These transitions are modeled using the SB, inverse dynamics model, and the reward model, ensuring that the resulting stitched trajectory remains coherent, without any discontinuities or unnatural transitions. In addition, further clarifications on the transitions at the stitching point are provided below, along with justifications and proof sketches for Theorem 1.

\section{Additional Justifications and Proof Sketches}
\subsection{Justifications of Lipschitz Assumption}
While the Lipschitz assumption may not hold universally across all clinical domains, we believe it is reasonable in our study for several reasons. First, physiological systems tend to behave smoothly at the macroscopic level, where small perturbations in patient states generally lead to proportionally small changes. This is due to homeostatic mechanisms and continuous biological processes. For instance, a slight increase in blood glucose levels rarely triggers abrupt physiological transitions, but rather induces gradual compensatory responses.
Second, in our framework, states typically represent aggregated clinical measurements such as vital signs and clinical values rather than microscopic biological states. At this level of abstraction, transitions are generally continuous and bounded. Third, the Lipschitz continuity assumption has been successfully employed in prior work on clinical applications. Specifically, prior work \cite{lipschitz_nips} has demonstrated the validity of this assumption for clinical time series modeling, achieving strong empirical results. Finally, in real-world medical practice, treatments are naturally constrained by safety considerations and physiological limitations. These constraints prevent abrupt or extreme changes in patient states, resulting in smooth and predictable treatment effects—exhibiting behavior similar to Lipschitz continuity. Therefore, by enforcing local smoothness through Lipschitz continuity, we can better preserve the clinical validity of stitched trajectories, ensuring they remain within a clinically plausible range and do not introduce any abrupt or unrealistic dynamics.

\subsection{Proof Sketches of Theorem 1}
\begin{proofsketch}
We analyze three cases for transitions in $\tau_{\text{stit}}$.

\textbf{Case 1: Transitions from $\tau_B$ (before stitching point).} For any transition $(\bar{s}, a, \bar{s}') := (s_j^B, a_j^B, s_{j+1}^B)$ where $j < t'$, this transition originally exists in $\tau_B \subset \mathcal{D}$. We can choose $(s, a, s') = (\bar{s}, a, \bar{s}')$, which gives $\|\bar{s} - s\| = 0 \leq \sqrt{2(1 - \delta)}$ and $\|\bar{s}' - s'\| = 0 \leq L\sqrt{2(1 - \delta)}$.

\textbf{Case 2: Transitions from $\tau_A$ (after stitching point).} 

\noindent
For any transition $(\bar{s}, a, \bar{s}') := (s_j^A, a_j^A, s_{j+1}^A)$ where $j > t$, this transition originally exists in $\tau_A \subset \mathcal{D}$. Again, taking $(s,a,s')=(\bar s,\bar a,\bar s')$ gives $\|\bar s - s\|=0$ and $\|\bar s' - s'\|=0$.

\textbf{Case 3: The transition process at the stitching point.} 

\noindent
The critical case is the transition $(\bar{s}, a, \bar{s}') := (s_{t'}^B, a_t^A, s_{t+1}^A)$ at the stitching point. From the cosine similarity condition in Equation \ref{eq:cosine_similarity_t}, we have:
\begin{equation}
\text{Sim}(s_t^A, s_{t'}^B) = \frac{\langle s_t^A, s_{t'}^B \rangle}{\|s_t^A\| \|s_{t'}^B\|} \geq \delta.
\end{equation}
Using the relationship between cosine similarity and Euclidean distance:
\begin{equation}
\|s_t^A - s_{t'}^B\|^2 = \|s_t^A\|^2 + \|s_{t'}^B\|^2 - 2\langle s_t^A, s_{t'}^B \rangle.
\end{equation}
Since $\text{Sim}(s_t^A, s_{t'}^B) \geq \delta$, we have $\langle s_t^A, s_{t'}^B \rangle \geq \delta \|s_t^A\| \|s_{t'}^B\|$. Assuming normalized states (or bounded norms), this yields:
\begin{equation}
\|s_t^A - s_{t'}^B\| \leq \sqrt{2(1 - \delta)}.
\end{equation}
Now, for the stitching transition, the state $\bar{s} = s_{t'}^B$, the action $a = a_t^A$ (the action originally taken from $s_t^A$), and the next state $\bar{s}' = s_{t+1}^A$. Consider the original transition $(s_t^A, a_t^A, s_{t+1}^A) \in \mathcal{D}$. We have:
\begin{equation}
\|\bar{s} - s_t^A\| = \|s_{t'}^B - s_t^A\| \leq \sqrt{2(1 - \delta)}.
\end{equation}
For the next state, using the Lipschitz property of $\mathcal{F}$:
\begin{equation}
\begin{split}
\|\bar{s}' - s_{t+1}^A\| &= \|\mathcal{F}(s_{t'}^B, a_t^A) - \mathcal{F}(s_t^A, a_t^A)\| \\
&\leq L \|s_{t'}^B - s_t^A\| \leq L\sqrt{2(1 - \delta)},
\end{split}
\end{equation}
where we use that $s_{t+1}^A = \mathcal{F}(s_t^A, a_t^A)$ by definition of the transition function.
For all transitions in $\tau_{\text{stit}}$, we have shown that there exists a corresponding transition in $\mathcal{D}$ within the specified bounds. The maximum deviation occurs at the stitching point and is bounded by $\mathcal{O}(L\sqrt{2(1 - \delta)})$. As $\delta \to 1$ (higher similarity threshold), this bound approaches 0, ensuring minimal distributional shift. This establishes that treatment stitching preserves the support of the original data distribution, thereby reducing OOD shifts.
\end{proofsketch}

\section{Additional Sensitivity and Ablation Study}
\subsection{Effect of the Reward Split Parameter $q$ in $\Phi$}
The reward percentile threshold $q$ in $\Phi_q(\mathcal{D})$ determines how trajectories are divided into high-reward ($\mathcal{D}_{\text{high}}$) and low-reward ($\mathcal{D}_{\text{low}}$) groups for the stitching process. In this work, we set $q = 50$ (i.e., the median) to bisect trajectories into high-reward and low-reward groups. We adopted this value due to its simple implementation and to ensure sufficient trajectories in both groups for effective priority sampling.

To assess the sensitivity of this parameter, we conducted additional experiments by increasing $q$ to 75 (upper quartile). As shown in Table \ref{appendix_table_q}, increasing $q$ to 75 led to modest performance improvements in the full data setting, as priority sampling could focus more heavily on higher-reward trajectories. However, in the restricted data setting, increasing $q$ provided no observable benefit. When the high-reward group becomes too small, the reduced diversity of trajectories limits effectiveness and can hinder learning process. 

\begin{table}[t]
\scriptsize
\centering
\renewcommand{\arraystretch}{1.5}
\setlength{\tabcolsep}{1.9pt}

\begin{tabular}{c|cccccccc|c}
\hline\hline
\textbf{Full Data} & \textbf{Env1} & \textbf{Env2} & \textbf{Env3} & \textbf{Env4} & \textbf{Env5} & \textbf{Env6} & \textbf{Env7} & \textbf{Env8} & \textbf{Mean} \\
\hline
Median ($q$ = 50) & \textbf{65.15} & 62.05 & \textbf{50.57} & 58.22 & 57.88 & \textbf{56.98} & 59.09 & \textbf{53.51} & 57.93 \\
Upper ($q$ = 75) & 64.92 & \textbf{62.28} & 50.43 & \textbf{58.71} & \textbf{58.12} & 56.85 & \textbf{59.33} & 53.42 & 58.01 \\
\hline
\end{tabular}

\vspace{6pt} 

\begin{tabular}{c|cccccccc|c}
\hline
\textbf{Restricted Data} & \textbf{Env1} & \textbf{Env2} & \textbf{Env3} & \textbf{Env4} & \textbf{Env5} & \textbf{Env6} & \textbf{Env7} & \textbf{Env8} & \textbf{Mean} \\
\hline
Median ($q$ = 50) & \textbf{42.22} & \textbf{43.21} & \textbf{30.29} & \textbf{33.63} & \textbf{34.78} & \textbf{34.68} & \textbf{36.03} & 30.19 & \textbf{35.63} \\
Upper ($q$ = 75) & 41.18 & 41.05 & 28.62 & 31.84 & 33.31 & 33.89 & 34.48 & \textbf{30.47} & 34.36 \\
\hline\hline
\end{tabular}

\caption{Effect of the reward split parameter ($q$) on the \textit{TreatStitch} under both full and restricted data settings.}
\label{appendix_table_q}
\end{table}

\begin{table}[t]
\scriptsize
\centering
\renewcommand{\arraystretch}{1.5}
\setlength{\tabcolsep}{1.9pt}
\begin{tabular}{c|cccccccc|c}
\hline\hline
\textbf{Method} & \textbf{Env1} & \textbf{Env2} & \textbf{Env3} & \textbf{Env4} & \textbf{Env5} & \textbf{Env6} & \textbf{Env7} & \textbf{Env8} & \textbf{Mean} \\
\hline
High-to-Low & 58.42 & 60.15 & 49.83 & 56.91 & 55.47 & 54.62 & 58.31 & \textbf{53.89} & 55.95 \\
Random Stitching & \textbf{65.28} & 59.03 & 50.44 & \textbf{60.87} & 56.21 & 56.45 & 55.82 & 52.64 & 57.09 \\
Low-to-High (Ours) & 65.15 & \textbf{62.05} & \textbf{50.57} & 58.22 & \textbf{57.88} & \textbf{56.98} & \textbf{59.09} & 53.51 & \textbf{57.93} \\
\hline\hline
\end{tabular}
\caption{Ablation study for comparing different data split strategies on EpiCare full data setting using CQL backbone.}
\label{appendix_table_split_ablation}
\end{table}

Overall, these results suggest that $q = 50$ provides a reasonable and robust balance across varying levels of data availability. In data-rich settings, increasing $q$ to 75 may yield additional benefits, whereas in data-scarce settings, the median split ($q = 50$) remains the reliable configuration. Exploring adaptive or dynamically adjusted values of $q$ based on dataset characteristics by leveraging hyperparameter optimization frameworks \cite{eswa_shin} would be an interesting direction for future study.

\subsection{Ablation Study on the Data Splitting Strategy}
In our framework, we strategically split trajectories into high-reward ($\mathcal{D}_{\text{high}}$) and low-reward ($\mathcal{D}_{\text{low}}$) groups and perform low-to-high stitching by combining initial segments from low-reward trajectories with later segments from high-reward trajectories. To assess the effectiveness of this design, we conducted an ablation study comparing three data-splitting strategies: low-to-high (our method), high-to-low, and random stitching without reward-based grouping. Specifically, the high-to-low strategy stitches high-reward trajectory prefixes with low-reward trajectory suffixes, producing trajectories that transition from beneficial to adverse outcomes. In contrast, the random stitching strategy constructs trajectories by randomly pairing prefixes and suffixes without relying on reward-based grouping.

As shown in Table~\ref{appendix_table_split_ablation}, both the high-to-low and random stitching strategies achieved comparable results, confirming the fundamental benefits of data augmentation through treatment stitching. Between the two, random stitching performs slightly better than high-to-low. In contrast, our low-to-high strategy delivers the best performance, outperforming both alternatives. The superior performance of our low-to-high strategy can be explained through reinforcement learning principles: agents aim to maximize cumulative returns by increasing the likelihood of actions that lead to high rewards. When the augmented dataset contains more trajectories that ultimately achieve high returns—as is the case with low-to-high stitching—the agent receives stronger learning signals about desirable long-term outcomes. This is especially important in offline RL, where the agent cannot interact with the environment and must rely entirely on the dataset to infer which action sequences lead to desirable outcomes. 

\section{Limitations and Future Study Directions}
In this work, we used cosine similarity to perform intermediate state similarity assessment due to its simplicity and solid empirical performance on benchmark datasets. However, we acknowledge that domain-informed similarity metrics tailored to clinical settings could further strengthen the validity of the stitching process. Although our theoretical analysis demonstrates that stitched trajectories remain within the bounds of real clinical data, additional clinically grounded evaluations—such as collaborating with physicians to conduct expert reviews of stitched trajectories—represent an important direction for future study. 

From a clinical perspective, several additional validation approaches would strengthen the framework's applicability to real-world clinical settings. Evaluating stitched trajectories against hard clinical constraints—such as medication dosage limits, contraindicated drug–drug interactions, and physiological safety thresholds—would help ensure that the augmented trajectories do not imply unsafe interventions. Moreover, presenting detailed case studies in which stitched trajectories align with established evidence-based treatment pathways would demonstrate that the framework captures clinically recognized therapeutic strategies. Finally, comparing stitched trajectories to formal clinical guidelines and treatment protocols would offer deeper insight into both the clinical validity and the practical utility.

Regarding our Schrödinger Bridge (SB) framework, enhancing the SB method by incorporating domain-specific constraints and expert-guided corrections represents a promising avenue for improvement. Currently, the SB method generates bridging states based purely on the optimal transport theory. Integrating clinical knowledge into this generation process, such as constraining bridging trajectories to respect known physiological relationships or incorporating physician feedback to guide the bridging process, could produce more clinically interpretable and reliable synthetic trajectories. Additionally, developing mechanisms to quantify and visualize the clinical plausibility of generated bridging states would help clinicians understand and trust the data augmentation process via treatment stitching.

Finally, clinical settings inherently require models that can generalize across diverse patient populations, varying disease presentations, and different healthcare contexts. Enhancing the generalization capability of our framework represents an important direction for future research. One promising approach is incorporating sharpness-aware minimization techniques, which have been shown to improve model robustness and generalization by seeking flatter minima in the loss landscape \cite{sam_shin}. 

\newpage
\clearpage
\section{Pseudo-Code}
In this section, we provide the pseudo-code for our proposed \textit{TreatStitch} framework. \textbf{Algorithm~\ref{alg:treatstitch}} outlines the main treatment stitching mechanism. This process constructs stitched trajectories by `stitching' together segments from two existing trajectories that share similar intermediate states, thereby preserving clinical validity while enhancing data diversity.

\begin{algorithm}[h]
\caption{Treatment Stitching (\textit{TreatStitch})}
\label{alg:treatstitch}

\KwIn{
    Offline dataset $\mathcal{D} = \{\tau_i\}_{i=1}^N$\\
    Reward percentile threshold $q$\\
    Similarity threshold $\delta$\\
    Number of new trajectories to generate $M$\\
    Temperature parameter $\alpha$
}
\KwOut{Augmented dataset $\mathcal{D}_{\text{aug}}$}

\BlankLine
Compute cumulative reward $R(\tau_i)$ for each trajectory $\tau_i \in \mathcal{D}$\;
Find the $q$-th percentile reward value $\Phi_q(\mathcal{D})$\;
$\mathcal{D}_{\text{high}} \leftarrow \{ \tau_i \in \mathcal{D} \mid R(\tau_i) \ge \Phi_q(\mathcal{D}) \}$\;
$\mathcal{D}_{\text{low}} \leftarrow \mathcal{D} \setminus \mathcal{D}_{\text{high}}$\;
$\mathcal{D}_{\text{aug}} \leftarrow \mathcal{D}$\;

\BlankLine
\For{$m = 1$ \KwTo $M$}{
    \Repeat{$\mathrm{Sim}(s_{\mathrm{stitch}}^A, s_{\mathrm{stitch}}^B) \ge \delta$}{

        \tcp{Priority Sampling}
        Sample $\tau_A \sim \mathcal{D}_{\text{high}}$ with probability 
        $p(\tau_A) \propto \exp(R(\tau_A)/\alpha)$\;
        Sample $\tau_B \sim \mathcal{D}_{\text{low}}$ with probability 
        $p(\tau_B) \propto \exp(-R(\tau_B)/\alpha)$\;

        \BlankLine
        \tcp{Intermediate State Similarity Assessment}

        \(
        \begin{aligned}
        (s_{\text{stitch}}^A, s_{\text{stitch}}^B)
        &\leftarrow \\
        &\operatorname*{arg\,max}_{s_t^A \in \tau_A,\; s_{t'}^B \in \tau_B}
         \mathrm{Sim}(s_t^A, s_{t'}^B)
        \end{aligned}
        \)\;

        $t_{\mathrm{stitch}} \leftarrow \operatorname{index}(s_{\mathrm{stitch}}^A, \tau_A)$\;
        $t'_{\mathrm{stitch}} \leftarrow \operatorname{index}(s_{\mathrm{stitch}}^B, \tau_B)$\;
    }

    \BlankLine
    \tcp{Stitched Trajectory}
    $\tau_{\text{prefix}} \leftarrow 
        \langle (s_k^B, a_k^B, r_k^B) \rangle_{k=0}^{t'_{\mathrm{stitch}}}$\;

    $\tau_{\text{suffix}} \leftarrow 
        \langle (s_k^A, a_k^A, r_k^A) \rangle_{k=t_{\mathrm{stitch}}+1}^{T_A}$\;

    $\tau_{\text{stit}} \leftarrow 
        \tau_{\text{prefix}} \oplus \tau_{\text{suffix}}$\;

    \BlankLine
    $\mathcal{D}_{\text{aug}} \leftarrow 
        \mathcal{D}_{\text{aug}} \cup \{\tau_{\text{stit}}\}$\;
}

\BlankLine
\Return{$\mathcal{D}_{\text{aug}}$}

\end{algorithm}

\textbf{Algorithm~2} describes the training procedure for bridging state generation, which is the core component of our extended framework, \textit{TreatStitch w/ SB}. In this scenario, the Schrödinger Bridge (SB) method is employed to generate smooth transitions (i.e., bridging states), further increasing stitching opportunities and enhancing model performance, especially under restricted data settings with sparse data.

We recall that the notations $\mathbf{x}$ and $s$ are used interchangeably to denote states. While $\mathbf{x}$ appears in the mathematical formulation of the SB method, $s$ refers to concrete states within trajectories (e.g., $s_t^C$, $s_{t'}^D$, or $\tilde{s}k$) in our \textit{TreatStitch} framework. However, both symbols refer to the same underlying entities in the state space. Within \textbf{Algorithm~2}, we describe the training process for generating bridging states using the SB-based generative model $G_\phi$. The goal is to learn a model capable of smoothly connecting two trajectory segments by generating a sequence of plausible bridging states.

\begin{algorithm}[h]
\caption{Bridging State Generation Training}
\KwIn{Generative model $G_\phi$, Start distribution $p_{\text{start}}$, Target distribution $p_{\text{target}}$, iterations $M$}
\KwOut{Trained model $G_\phi$}

Initialize generative model $G_\phi$\
\BlankLine
\For{$m = 1$ \KwTo $M$}{
  \(t_i \sim \mathcal{U}(0, 1)\)\\
  Sample $\mathbf{x}_{\text{start}} \sim p_{\text{start}}$, \ $\boldsymbol{\epsilon}^{\text{fwd}}_i \sim \mathcal{N}(\mathbf{0}, \mathbf{I})$\
  $\tilde{\mathbf{x}}^{\text{fwd}}_i = \mathbf{x}_{\text{start},i} + \sigma \sqrt{t_i}\ \boldsymbol{\epsilon}^{\text{fwd}}_i$\

  Sample $\mathbf{x}_{\text{target}} \sim p_{\text{target}}$, \ $\boldsymbol{\epsilon}^{\text{bwd}}_i \sim \mathcal{N}(\mathbf{0}, \mathbf{I})$\
  $\tilde{\mathbf{x}}^{\text{bwd}}_i = \mathbf{x}_{\text{target},i} + \sigma \sqrt{1-t_i}\, \boldsymbol{\epsilon}^{\text{bwd}}_i$\

  Compute $\mathcal{L}_{\text{fwd}}(\phi)$ using Eq.~\ref{fwd_loss}\\
  Compute $\mathcal{L}_{\text{bwd}}(\phi)$ using Eq.~\ref{bwd_loss}\\

  Compute $\mathcal{L}_{\text{total}}(\phi) = \mathcal{L}_{\text{fwd}}(\phi) + \mathcal{L}_{\text{bwd}}(\phi)$
  
  Update model $G_\phi$ with gradient $\nabla_{\phi}\mathcal{L}(\phi).$  

}

\end{algorithm}
\end{document}